\documentclass{article}

\usepackage{arxiv}

\usepackage{hyperref}

\usepackage{babel}
\DeclareMathAlphabet\mathbfcal{OMS}{cmsy}{b}{n}

\usepackage{amsmath,amsthm,amssymb,babel,amsfonts,graphics, mathtools}

\usepackage{booktabs}

\usepackage{algorithm}
\usepackage{algpseudocode}
\usepackage{enumitem}
\usepackage{dirtytalk}

\usepackage{caption}
\usepackage{subcaption}
\usepackage{enumitem}
\DeclareFontShape{OT1}{cmr}{bx}{sc}{<->cmr10}{}
\usepackage{amsmath,amssymb,amsfonts,latexsym,cancel}
\usepackage{rawfonts}
\usepackage{pictexwd}
\usepackage{tikz}

\usepackage{tikz-cd}
\usepackage{mathrsfs}
\usepackage{multirow}
\usepackage{array} 

\usepackage{pgfplots}

\pgfplotsset{compat=newest}
\pgfplotsset{plot coordinates/math parser=false}
\newlength\figureheight
\newlength\figurewidth

\usepackage{color}
\usepackage{epsfig}

\newcommand{\R}{\mathbb{R}}

\def\1{\raisebox{2pt}{\rm{$\chi$}}}

\theoremstyle{plain}
\newtheorem{theorem}{THEOREM}[section]
\newtheorem{corollary}[theorem]{COROLLARY}

\theoremstyle{definition}
\newtheorem{definition}[theorem]{DEFINITION}

\theoremstyle{definition}

\newtheorem{lemma}[theorem]{LEMMA}

\theoremstyle{remark}
\newtheorem{remark}[theorem]{REMARK}

\theoremstyle{definition}

\theoremstyle{definition}

\theoremstyle{remark}

\numberwithin{equation}{section}

\DeclareUnicodeCharacter{2212}{_}
\DeclareUnicodeCharacter{0014}{-}

\title{Deep Network Trainability via Persistent Subspace Orthogonality}
\author{ 
Alex Massucco\footnotemark[1]\\
	Department of Applied Mathematics\\
	and Theoretical Physics,\\
	University of Cambridge, Cambridge, UK\\
	\texttt{am3270@cam.ac.uk}\\
\And
Davide Murari\footnotemark[1]\\
	Department of Applied Mathematics\\
	and Theoretical Physics,\\
	University of Cambridge, Cambridge, UK\\
	\texttt{dm2011@cam.ac.uk}
\And
Carola-Bibiane Schönlieb\\
	Department of Applied Mathematics\\
	and Theoretical Physics,\\
	University of Cambridge, Cambridge, UK\\
	\texttt{cbs31@cam.ac.uk}
}
\date{}

\begin{document}

\maketitle
\footnotetext[1]{These authors contributed equally to this work}

\begin{abstract}
Training neural networks via backpropagation is often hindered by vanishing or exploding gradients. In this work, we design architectures that mitigate these issues by analyzing and controlling the network Jacobian. We first provide a unified characterization for a class of networks with orthogonal Jacobian including known architectures and yielding new trainable designs. We then introduce the relaxed notion of \emph{persistent subspace orthogonality}. This applies to a broader class of networks whose Jacobians are isometries only on a non-trivial subspace. We propose practical mechanisms to enforce this condition and empirically show that it is necessary to sufficiently preserve the gradient norms during backpropagation, enabling the training of very deep networks. We support our theory with extensive experiments.
\end{abstract}

\section{Introduction}\label{se:introduction}

Neural networks can approximate unknown functions accurately, and their expressivity typically increases with depth \cite{poggio2017and,yarotsky2017error}. However, deeper networks are harder to train with the backpropagation algorithm due to instability in gradient propagation \cite{hochreiter2001gradient}. Residual neural networks (ResNets) were introduced in \cite{KaimingHe2015} to improve the trainability of deep models by adding skip connections, which facilitate signal propagation and, under suitable weight initialization, mitigate vanishing and exploding gradients \cite{bengio1994,glorot2010}.

In \cite{saxe2014}, the authors analyze the learning dynamics of deep linear networks and show that orthogonal weight initialization yields depth-independent learning times. They introduce the term \emph{dynamical isometry}, i.e., the regime where the singular values of the network Jacobian are tightly concentrated around one at initialization. This notion has enabled training extremely deep convolutional networks even without residual connections \cite{xiao2018}. In \cite{pennington2017}, the authors study how Jacobian singular values depend on the activation function and conclude that feedforward $\mathrm{ReLU}$ networks cannot achieve dynamical isometry.


\begin{figure*}
    \centering
    \includegraphics[width=0.9\linewidth]{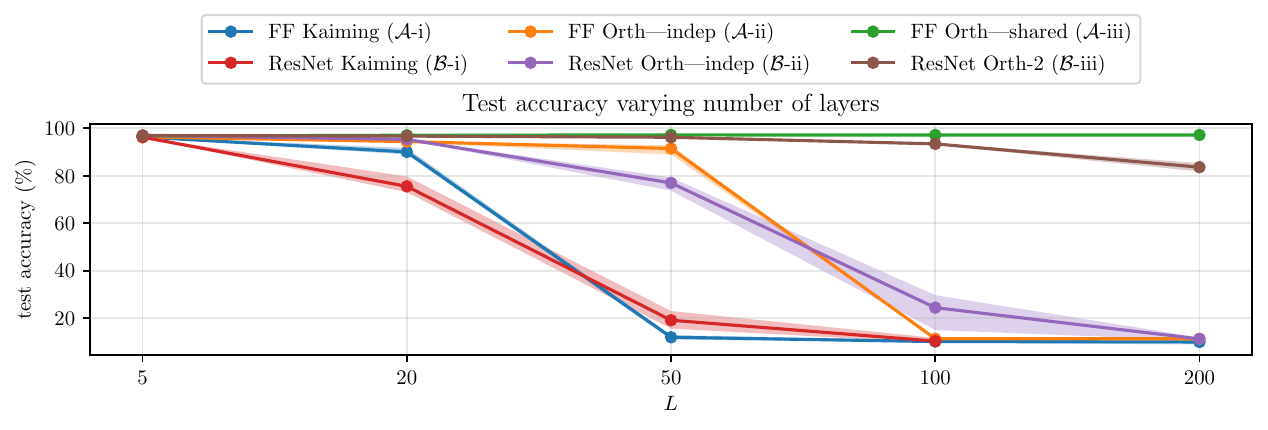}
    \caption{Test accuracy over 5 runs for a feedforward $\mathrm{ReLU}$-network (type \eqref{eq:type_A}) and a residual $\mathrm{ReLU}$-network (type \eqref{eq:type_B}) with different initializations. We consider $L\in\{5,20,50,100, 200\}$ layers. The plotted results are the mean accuracies, and the confidence interval is determined from the minimum and maximum absolute deviations from the mean.}
    \label{fig:motivation}
\end{figure*}

To motivate a relaxed notion of dynamical isometry, we study MNIST classification at depths $L\in\{5,20,50,100, 200\}$ with width $d=64$. Each hidden layer is either a feedforward block
\begin{equation}\tag{$\mathcal{A}$}\label{eq:type_A}
F_1(x)=A^\top \mathrm{LeakyReLU}_\alpha(Bx+b)\in\mathbb{R}^d,
\end{equation}
or a residual block
\begin{equation}\tag{$\mathcal{B}$}\label{eq:type_B}
F_2(x)=x+A^\top \mathrm{LeakyReLU}_\alpha(Bx+b)\in\mathbb{R}^d,
\end{equation}
where $\mathrm{LeakyReLU}_\alpha(s)=\max\{\alpha s,s\}$, $A,B\in\mathbb{R}^{d\times d}$ and $b\in\mathbb{R}^d$. Biases are initialized to zero. For each block type, we compare three initialization schemes: (i) Kaiming normal for $A,B$; (ii) independent random orthogonal $A,B$ resampled in each layer; (iii) a structured orthogonal initialization. For type \eqref{eq:type_A}, we sample a single orthogonal matrix $Q$ once and set $A=B=Q$ in every layer (shared across depth). For type \eqref{eq:type_B}, in each layer we sample an orthogonal matrix $B$ (which may vary with depth) and initialize $A$ to  $- 2B$. We repeat each configuration five times, focusing on the negative slope $\alpha=0$, and report the mean test accuracy in Figure~\ref{fig:motivation}. Further details and experiments are provided in Section~\ref{se:experiments}. These experiments show that the only configurations that reliably train across various depths are (\ref{eq:type_A}-iii) and (\ref{eq:type_B}-iii). The latter satisfies dynamical isometry (Section~\ref{se:theory}), whereas the former satisfies the relaxed notion of \emph{persistent subspace orthogonality} we introduce in Section~\ref{se:PSO}. 

\subsection*{Main contributions}
We first identify structural conditions under which perfect Jacobian orthogonality is achievable, emphasizing the coupling required between the activation function and the skip connection. We then show that these conditions can be relaxed and introduce the new concept of \emph{persistent subspace orthogonality} that requires the Jacobian to be an isometry only on a suitable subspace, leading to an initialization strategy that enables training deep feedforward $\mathrm{ReLU}$-networks.

Finally, we present numerical experiments that support the theory and demonstrate that the proposed architectures remain trainable at substantial depth.

\subsection*{Outline of the paper}

The paper is organized as follows. Section~\ref{se:theory} presents a general framework for neural networks under Jacobian orthogonality constraints and derives several instances that recover both standard and novel architectures. Section~\ref{se:PSO} formalizes \emph{persistent subspace orthogonality}, shows why a shared isometric subspace is necessary, and explains how this relaxes dynamical isometry. Section~\ref{se:experiments} and Section~\ref{app:ablations} provide empirical validation of our theoretical derivations discussing the numerical experiments and several ablation tests. We conclude in Section~\ref{se:discussion} with a summary and further directions.

\subsection*{Notation}
We denote the set of $n\times n$ orthogonal matrices as $\mathcal{O}(n) =\{A\in\mathbb{R}^{n\times n}:\,A^\top A=AA^\top=I_n\}$, where $I_n\in\mathbb{R}^{n\times n}$ is the identity matrix, and write $1_n,0_n\in\mathbb{R}^n$ for the vectors of ones and zeros, respectively. $0_{n,m}\in\mathbb{R}^{n\times m}$ denotes the zero matrix. Depending on the context, \textit{a.e.} stands for \textit{almost every} or \textit{almost everywhere}. For an a.e.\, differentiable vector field $F:\mathbb{R}^n\to\mathbb{R}^n$, we denote by $F'(x)\in\mathbb{R}^{n\times n}$ its Jacobian at $x$ when defined. We use $\sigma(\cdot\,;\alpha,\beta):\mathbb{R}\to\mathbb{R}$ to indicate a Lipschitz function such that $\sigma'(t)\in\{\alpha,\beta\}$ for a.e. $t$, where $\alpha,\beta\in\mathbb{R}$, and write $\mathrm{LeakyReLU}_{\alpha}(s)=\max\{\alpha s,s\}$, $\alpha\in [0,1)$ when referring to the LeakyReLU function with negative slope $\alpha$. For $\Omega_i\subset\mathbb{R}^n$, $1_{\Omega_i}:\mathbb{R}^n\to\mathbb{R}$ denotes the indicator function, i.e., $1_{\Omega_i}(x)=1$ if $x\in\Omega_i$ and $0$ otherwise. We write $\overline{\Omega}_i$ for the closure of $\Omega_i$.

\section{Related works}

\paragraph{Vanishing and exploding gradients.} Skip connections and dynamical isometry are not the only mechanisms used to mitigate vanishing and exploding gradients. A first line of work enforces structural constraints on the Jacobian through architectural design, as in Hamiltonian/symplectic networks \cite{maslovskaya2024symplectic,galimberti2023hamiltonian,haber2017stable} and reversible networks \cite{unterthiner2016understanding}. A second family of methods improves gradient flow through gating, e.g., LSTMs \cite{hochreiter1997long} and Highway networks \cite{srivastava2015highway}. Normalization layers, including batch, layer, and group normalization, stabilize activation scales across depth and can improve gradient propagation \cite{ioffe2015batch,ba2016layer,wu2018group}. Since activation functions strongly influence gradient attenuation, substantial effort has gone into well-behaved nonlinearities such as SeLU \cite{klambauer2017self} and (Leaky)$\mathrm{ReLU}$ variants \cite{xu2015empirical,maas2013rectifier}; similarly, GroupSort \cite{anil2019sorting} yields an orthogonal Jacobian a.e.\, but is not applied entrywise. Finally, exploding gradients are often addressed in practice via gradient clipping \cite{pascanu2013difficulty}.

\paragraph{Jacobian orthogonality and Lipschitz constraints.} For linear layers, orthogonal weights yield an orthogonal Jacobian. More generally, orthogonality-inspired constraints are used to promote stable signal propagation and to control the network Lipschitz constant \cite{wang2020orthogonal,singla2021skew,trockman2021orthogonalizing}. Lipschitz-controlled networks have been leveraged for adversarial robustness \cite{tsuzuku2018lipschitz,sherry2024designing,trockman2021orthogonalizing,prach20241,meunier2022dynamical}, for generative modeling \cite{gulrajani2017improved,miyato2018spectral,lunz2018adversarial}, and for provably convergent or stable learned procedures in inverse problems \cite{sherry2024designing,ryu2019plug,hertrich2021convolutional}.

\paragraph{Functions with orthogonal Jacobian a.e.\,.} Orthogonality of the Jacobian (often studied as a form of local isometry) also appears in analysis and geometry. We refer to \cite{dacorogna2008lipschitz,chambolle2007piecewise,dacorogna2010functions,caboussat2019numerical}, which study such maps in origami and folding via partial differential inclusions, with further applications in plate theories \cite{kirchheim2003,conti2008confining}. These works are general and largely non-constructive. In contrast, we provide a constructive, self-contained analysis of maps resembling standard neural networks and compatible with efficient implementation.

\paragraph{Piecewise-linear activation functions.} Our constructions require piecewise-linear activation functions with controlled slopes. Beyond standard $\mathrm{ReLU}$ and $\mathrm{Leaky\mathrm{ReLU}}$, the literature includes learnable variants such as $\mathrm{PReLU}$ \cite{he2015delving}, where the negative-slope parameter is trained, as well as richer parameterizations with more flexibility \cite{zhou2021learning,jin2016deep,bohra2020learning}.

\section{Theoretical characterization: exact Jacobian orthogonality}\label{se:theory}
We now provide a characterization of a broad class of vector fields whose Jacobian matrix is orthogonal for a.e.\, input $x\in\mathbb{R}^n$. These maps provide the skeleton of the more general class of networks that we discuss in Section \ref{se:PSO} satisfying our newly introduced concept of \textit{persistent subspace orthogonality}. The vector fields we consider in this section are defined based on partitioning the input space $\mathbb{R}^n$ into open and connected sets. Lemma \ref{lemma:char_orth_field} shows that, when the vector field is sufficiently regular, it has an orthogonal Jacobian matrix a.e.\, if and only if it is piecewise affine. We then focus on a family of parametric vector fields resembling commonly implemented neural networks. In Theorem \ref{thm:char_RtoR_fields}, we show how to constrain those neural networks to have an orthogonal Jacobian matrix for a.e.\, input.

The vector fields considered in this section may exhibit jump discontinuities. Then, given the growing interest in networks with discontinuities (e.g., \cite{kong2025expressivity,della2023discontinuous,yin2018understanding}), we emphasize that our results are also practically relevant to this research line.

\begin{definition}\label{ass:partition}
Let $\Omega\subseteq\mathbb{R}^n$ be a closed connected set and $N\in\mathbb{N}$. We say that $\{\Omega_i\subseteq\R^n\mid\,i = 1,\cdots,N\}$ belongs to $\mathcal{P}(\Omega)$ if:
\begin{enumerate}
\item $\Omega_1,\cdots,\Omega_N \subseteq \Omega$ are disjoint open and connected sets,
\item $ \Omega = \bigcup_{i = 1}^{N} \overline{\Omega}_i$,
\item $\partial {\overline{\Omega}_1},\cdots,\partial {\overline{\Omega}_N}$ are Lipschitz boundaries with zero $n$-dimensional Lebesgue measure.
\end{enumerate}
\end{definition}

We now present a fundamental definition, which is given in Definition 2.2 of \cite{dacorogna2008lipschitz}.
\begin{definition}[Piecewise $C^1$ map]
Let $F:\Omega\subseteq\mathbb{R}^n\to\mathbb{R}^n$ be Lipschitz continuous. Let $\Sigma\subset\Omega$ be the set of points where $F$ is not differentiable. $F$ is piecewise $C^1$ if (i) $\Sigma$ is closed in $\Omega$, (ii) $F$ is continuously differentiable on the connected components of $\Omega\setminus\Sigma$, and (iii) for any compact set $K\subset\Omega$, the number of connected components of $\Omega\setminus\Sigma$ intersecting $K$ is finite.
\end{definition}
\begin{definition}[Piecewise affine]
Let $F:\Omega\subseteq\mathbb{R}^n\to\mathbb{R}^n$ be Lipschitz continuous and piecewise $C^1$. We say that $F$ is piecewise affine, and we write $F \in \mathscr{A}$, if it is affine on every connected component of $\Omega\setminus\Sigma$, where $\Sigma$ is the non-differentiability set of $F$.
\end{definition}
\begin{lemma}\label{lemma:char_orth_field}
Let $\{\Omega_i\subseteq\mathbb{R}^n|\,i=1,\cdots,N\}$ be a collection of sets in $\mathcal{P}(\Omega)$, with $\Omega\subseteq\mathbb{R}^n$ connected. Consider a vector field $F:\mathbb{R}^n\to\mathbb{R}^n$ whose restriction to $\Omega_i$, which we call $F_i:\Omega_i\to\mathbb{R}^n$, is piecewise $C^1$ for every $i=1,\cdots,N$. Then, $F'(x) \in \mathcal{O}(n)$ for a.e.\, $x\in\Omega$ if and only if $F_1,\cdots,F_N$ are piecewise affine.
\end{lemma}
We omit the proof of this lemma, which can be found in Lemma 4.1 of \cite{dacorogna2008lipschitz}.

Let us now focus on maps resembling commonly implemented neural network layers. We consider a particular class of functions $F:\Omega\subseteq\mathbb{R}^n\to\mathbb{R}^n$ of the form
\begin{equation}\label{eq:focusMap}
F(x) = \sum_{i=1}^N 1_{\Omega_i}(x)\left( g_i(x) + d_iA^\top \sigma_i(Bx+b)\right)
\end{equation}
where $\{\Omega_1,\cdots,\Omega_N\}$ is in $\mathcal{P}(\Omega)$, $A,B\in\mathbb{R}^{n\times n}$, $b\in\mathbb{R}^n$, and $d_i\in\mathbb{R}\setminus \{0\}$. We assume that, for every $i\in\{1,\cdots,N\}$, $g_i(x)=\ell_i x + c_i$ with $\ell_i,c_i\in\mathbb{R}$, and $\sigma_i:\R\to\R$ is a piecewise $C^1$ Lipschitz continuous scalar function applied entrywise. We now characterize how $A$, $B$, and the activation functions $\sigma_1,\cdots,\sigma_N:\mathbb{R}\to\mathbb{R}$ need to be chosen for $F$ to have an orthogonal Jacobian a.e., whatever the partition $\{\Omega_1,\cdots,\Omega_N\}$ under consideration.

We remark that if, for example, $N=1$, $\Omega=\mathbb{R}^n$, $g_1(x)=x$, 
and $\sigma_1(x)=\mathrm{ReLU}(x)$, we recover a rather standard ResNet layer $F(x)=x+A^\top \mathrm{ReLU}(Bx+b)$, whereas if $N=1$, $\Omega=\mathbb{R}^n$, $g_1(x)=0$, and $\sigma_1(x)=\mathrm{ReLU}(x)$ we recover the feedforward network layer $F(x)=A^\top \mathrm{ReLU}(Bx+b)$.

We now define a set of functions that are fundamental to our derivation and then present the first main result of the paper.

\begin{definition}
We denote with $\mathcal{L}(\alpha,\beta)$ the set of scalar piecewise affine and continuous functions with slopes $\alpha,\beta\in\mathbb{R}$:
\[
\begin{split}
\mathcal{L}(\alpha,\beta) = \{&\sigma:\mathbb{R}\to\mathbb{R}:\sigma'(x)\in\{\alpha,\beta\}\,\text{ a.e.}\}\cap\mathscr{A}\cap C,
\end{split}
\]
where $C$ is the space of continuous functions.
\end{definition}
\begin{theorem}\label{thm:char_RtoR_fields}
Let $N\in\mathbb{N}$ and $\{\Omega_1,\cdots,\Omega_N\}$ in $\mathcal{P}(\mathbb{R}^n)$ be arbitrary. Consider $F:\R^n\to\R^n$ defined as in \eqref{eq:focusMap}, with $\sigma_i:\mathbb{R}\to\mathbb{R}$ Lipschitz continuous and piecewise $C^1$, and $g_i(x)=\ell_ix+c_i$, $\ell_i,c_i\in\mathbb{R}$, for every $i=1,\cdots,N$. Then, $F'(x)$ is orthogonal for a.e.\, $x\in\mathbb{R}^n$ if and only if either
\begin{enumerate}
    \item\label{ci} $A,B\in\mathcal{O}(n)$, $g_i(x)=c_i\in\mathbb{R}$, and $\sigma_i\in\mathcal{L}\left(-\frac{1}{d_i},\frac{1}{d_i}\right)$, for any $i=1,\cdots,N$, or 
    \item\label{cii}$A=B\in\mathcal{O}(n)$, $g_i(x)=\ell_i x+ c_i$, with $\ell_i,c_i\in\mathbb{R}$, and $\sigma_i\in\mathcal{L}\left(\frac{1-\ell_i}{d_i},-\frac{1+\ell_i}{d_i}\right)$, for any $i=1,\cdots,N$.
\end{enumerate}
\end{theorem}
See Appendix \ref{app:proofThm1} for the proof of Theorem \ref{thm:char_RtoR_fields}. Our proof is by exhaustion, in that we show that the constraints provided by the assumptions of Theorem \ref{thm:char_RtoR_fields} restrict the allowed options to cases \ref{ci} and \ref{cii}.
\begin{remark}
Assuming that $g_1,\cdots,g_N:\R\to\R$ are affine is not restrictive. Lemma \ref{lemma:char_orth_field} ensures that if $g_i:\mathbb{R}\to\mathbb{R}$ is piecewise $C^1$ and Lipschitz as $\sigma_i$, and $F$ has a Jacobian matrix which is orthogonal a.e., then $F_i(x)=g_i(x)+d_iA^\top \sigma_i(Bx+b)$ must be piecewise affine. Therefore, given that the partition $\{\Omega_1,\cdots,\Omega_N\}$ and the number of its elements $N\in\mathbb{N}$ are arbitrary, we can refine the partition so that on every set $\Omega_i$ the function $g_i$ is affine.
\end{remark}
We remark that when we add a scalar to a vector, as in $\ell_ix+c_i$, we use a more compact notation for $\ell_ix+c_i1_n$. It is also clear that maps of the form $F_{i,v_i}(x)=F_i(x)+v_i$, with $v_i\in\mathbb{R}^n$ and $F_i$ as in Theorem \ref{thm:char_RtoR_fields}, still have an orthogonal Jacobian. Furthermore, since the product of orthogonal matrices is orthogonal, one could get networks with Jacobians that are orthogonal for a.e.\, $x$ by composing the above layers with global isometries, i.e., maps of the form $x\mapsto Rx+b$ with $R\in\mathcal{O}(n)$ and $b\in\mathbb{R}^n$. This operation also allows encoding a generalization of the residual layers of case \ref{cii}, as formalized in the next corollary.
\begin{corollary}[Corollary to Theorem \ref{thm:char_RtoR_fields}]\label{co:corollary}
Let $N\in\mathbb{N}$ and $\{\Omega_1,\cdots,\Omega_N\}$ in $\mathcal{P}(\mathbb{R}^n)$ be arbitrary. Consider $G:\R^n\to\R^n$ defined as
\[
G(x) = \sum_{i=1}^N1_{\Omega_i}(x)\left(\ell_i Ox + c_i + d_iA^\top \sigma_i(Bx+b)\right),
\]
$\sigma_i:\mathbb{R}\to\mathbb{R}$ Lipschitz continuous and piecewise $C^1$, $c_i,d_i\in\mathbb{R}$, $\ell_i\in\mathbb{R}\setminus\{0\}$, for every $i=1,\cdots,N$, and $A,B,O\in\mathcal{O}(n)$. Then, $G'(x)$ is orthogonal for a.e.\, $x\in\mathbb{R}^n$ if and only if $\sigma_i\in\mathcal{L}\left(\frac{1-\ell_i}{d_i},-\frac{1+\ell_i}{d_i}\right)$ and $A=BO^\top$, i.e., if and only if $G(x)=OF(x)$ with $F$ as in case \ref{cii}.
\end{corollary}
We prove this result in Appendix \ref{app:proofCorollary}. We now derive three particular maps arising from \eqref{eq:focusMap}, with the constraints given by Theorem \ref{thm:char_RtoR_fields}.

\begin{figure}
    \centering
    \begin{subfigure}{0.4\textwidth}
    \includegraphics[width=6cm]{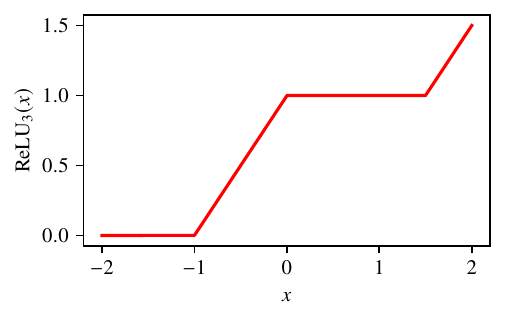}
    \caption{ $\mathrm{ReLU}_3\in\mathcal{L}(0,1)$.}
    \label{fig:ReLUk}
    \end{subfigure}
    \begin{subfigure}{0.4\textwidth}
     \includegraphics[width=6cm]{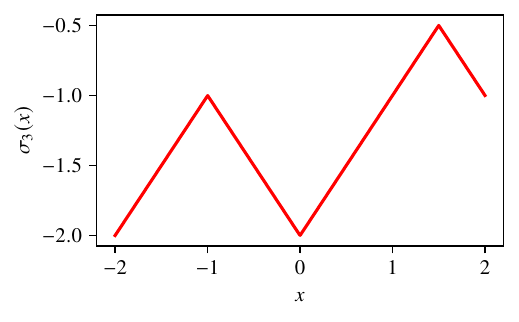}
      \caption{$\sigma_3\in\mathcal{L}(-1,1)$.}
      \label{fig:absk}
    \end{subfigure}
    \caption{Plot of two piecewise affine Lipschitz continuous scalar functions.}
\end{figure}

\paragraph{Modified ResNet layer} The first choice is given by
\begin{equation}\label{eq:ReLUk}
F(x) = x - 2B^\top \mathrm{ReLU}_k(B x + b),
\end{equation}
where, setting $x_1<x_2<\cdots<x_k$,
\begin{equation}\label{eq:stair}
\mathrm{ReLU}_k(x) = \sum_{i=1}^k (-1)^{i-1}\mathrm{ReLU}(x-x_i),
\end{equation}
is a piecewise affine Lipschitz-continuous function with slopes either $0$ or $1$, i.e., $\mathrm{ReLU}_k\in\mathcal{L}(0,1)$. 
The case $k=1$ with $x_1=0$ coincides with the usual $\mathrm{ReLU}(x)=\max\{0,x\}$. The case $k=2$ allows us to represent the activation $\mathrm{HardTanh}$. We include the plot of $\mathrm{ReLU}_3$, with $x_1=-1$, $x_2=0$, and $x_3=1.5$, in Figure \ref{fig:ReLUk}.
The $F$ in \eqref{eq:ReLUk} corresponds to case \ref{cii} with $N=1$, $l_1=1$, $c_1=0$, and $d_1 = -2$ so that $\frac{1- l_1}{d_1} = 0$, and $-\frac{1+l_1}{d_1}=1$. The layer with $k=1$ appears in \cite{meunier2022dynamical,prach20241,sherry2024designing,murari2025approximation}, and we test it in Section \ref{se:experiments}.

\paragraph{Modified feedforward network} We now consider a case without skip connection, i.e., $g(x)=0$. Fixing $N=1$ and $d_1=1$, in \ref{ci}, and choosing $\sigma(x)=|x|$, we recover the layer $F(x) = A^\top |Bx+b|$, where $|Bx+b|_i = |(Bx+b)_i|$. $|\cdot |$ can also be replaced by $\sigma_k\in\mathcal{L}(-1,1)$ defined as
\begin{equation}\label{eq:generalAbs}
\sigma_k(x) = x - 2\mathrm{ReLU}_k(x),
\end{equation}
with $\mathrm{ReLU}_k$ as in \eqref{eq:stair}.
We plot in Figure \ref{fig:absk} the function $\sigma_3$ with the same choices for $x_1,x_2,x_3$ as for $\mathrm{ReLU}_3$. The lack of monotonicity allows $\sigma_k$ to generate feedforward layers with $\|F'(x)\|_2=1$ a.e., which would not be possible for monotonic $1$-Lipschitz activations, see Theorem 1 of \cite{anil2019sorting}. The choice $k=1$ is considered in the experiments of Section \ref{se:experiments}.

We test these layers with activation functions $\mathrm{ReLU}_3$ and $\sigma_3$ in Section \ref{app:ablations}.

\paragraph{Discontinuous skip connection} Finally, we exploit the degree of freedom provided by the underlying partition and derive a discontinuous map presenting a gating mechanism. Let $\Omega_1 = \{x\in\R^n:\,\,a^\top x < 0\}$ and $\Omega_2 = \{x\in\R^n:\,\,a^\top x > 0\}$, where $a\in\R^n$ could be a learnable vector. We then define the skip-connections as
\[
g(x) = -1_{\Omega_1}(x) x + 1_{\Omega_2}(x) x,
\]
meaning that $l_1 = -1$, $l_2=1$, and $c_1=c_2=0$. A compatible choice of activation functions leads to
\[
\rho(x) = 2(1_{\Omega_1}(x) \mathrm{ReLU}_k(x) - 1_{\Omega_2}(x) \mathrm{ReLU}_k(x)),
\]
where $k\in\mathbb{N}$, and we have fixed $d_1=d_2=1$ for simplicity. We thus obtain the network layer
\begin{align*}
F(x) & = g(x)+B^\top \rho(Bx+b)\\
& = 1_{\Omega_1}(x) \left(-x + 2B^\top\mathrm{ReLU}_k(Bx+b)\right)+ 1_{\Omega_2}(x) \left(x - 2B^\top\mathrm{ReLU}_k(Bx+b)\right).
\end{align*}

Setting $s(x)=\mathrm{sign}(a^\top x)$, we can rewrite the above layer as 
\[
F(x) = s(x) (x - 2 B^\top \mathrm{ReLU}_k(Bx+b)),
\]
where $\mathrm{ReLU}_k$ is defined as in \eqref{eq:stair} for a suitable choice of breaking points $x_1,\cdots,x_k$. The resulting map, up to a sign change, is the same vector field we constructed in \eqref{eq:ReLUk}, however, in this case it is not globally continuous and the multiplying factor $s(x)$ can be interpreted as a gating mechanism, similar to those of Highway Networks \cite{srivastava2015highway} or LTSMs \cite{hochreiter1997long}.

\subsection{Dependence on the partition} The theory developed so far is independent of the underlying partition $\{\Omega_1, \cdots, \Omega_N\}$, but instead only relies on the selection of $N$ and the scalars $\{(l_i,c_i)\,|\,i=1,\cdots,N\}$. In the remainder of this section, we reverse this perspective and investigate the role of the chosen partition.

Let $\Omega\subset\mathbb{R}^n$ be a compact and connected set, and consider a collection of subsets $\{\Omega_i\subseteq\Omega|\,i=1,\cdots,N\}$ in $\mathcal{P}(\Omega)$. If $d_1=\cdots=d_N = 1$, any $F$ as in \eqref{cii} can be rewritten as
\begin{align}
&\mathcal{F}_\Omega(x; \widetilde{m}, \widetilde{q})=F(x) = \widetilde{m}(x)x + \widetilde{q}(x) + 1_{\Omega}(x)B^\top \sigma(Bx+b;\,\,1-\widetilde{m}(x),-1-\widetilde{m}(x)),\label{eq:pwcontinuous}
\end{align}
with
\begin{align}
\widetilde{m}(x)&=\sum_{i=1}^N 1_{\Omega_i}(x)\ell_i,\,\,\widetilde{q}(x)=\sum_{i=1}^N 1_{\Omega_i}(x) c_i. \nonumber
\end{align}
The primary advantage of this new formulation is that it embeds the underlying partition within the piecewise-constant functions $\tilde{m}$ and $\tilde{q}$. We remark that the notation $\sigma(f(x)\,;m_1(x),m_2(x))$ involves a slight abuse in notation since it refers to a function which at every $x$ has either slope $m_1(x)$ or $m_2(x)$. However, it can also be non-affine.

\begin{theorem}\label{thm:density}
Let us consider a compact and connected set $\Omega\subset\R^n$, $B\in\mathcal{O}(n)$, $b\in\R^n$, and $\varepsilon>0$. Then, for any pair of continuous functions $m,q:\Omega\to\R$, there exists $N\in\mathbb{N}$, a finite collection of subsets $\{\Omega_i\subseteq\Omega|\,i=1,\cdots,N\}$ in $\mathcal{P}(\Omega)$, and a choice of $2N$ scalars $m_1,\cdots,m_N,q_1,\cdots,q_N\in\R$ such that
\[
\max_{x\in\Omega}\left\|\mathcal{F}_\Omega(x; m, q) - \mathcal{F}_\Omega(x; \widetilde{m}, \widetilde{q})\right\|_{\infty} < \varepsilon,
\]
where $\mathcal{F}_\Omega(x; \cdot, \cdot)$ is given by \eqref{eq:pwcontinuous} and $\widetilde{m}(x) = \sum_{i=1}^N 1_{\Omega_i}(x)m_i$, $\widetilde{q}(x) = \sum_{i=1}^N 1_{\Omega_i}(x)q_i$.
\end{theorem}
See Appendix \ref{app:limitLayers} for the proof. There, we expand on this construction and provide more details on how close these generalized maps are to having an orthogonal Jacobian, as well as on the singular values of their Jacobians. We also include numerical experiments investigating their trainability.

\section{Persistent subspace orthogonality}\label{se:PSO}
In this section, working as before at the layer level, we present a generalized notion of orthogonality and demonstrate its effect on network training. We first introduce the concept of \textit{subspace orthogonality} and explain why it is insufficient to ensure trainability. We address this issue by identifying the necessity of a shared subspace where the desired orthogonality holds, i.e., the \textit{persistent subspace orthogonality}.

\paragraph{Subspace orthogonality}
\begin{definition}[Subspace orthogonality]
The Lipschitz-continuous function $F:\mathbb{R}^n\to\mathbb{R}^n$ satisfies subspace orthogonality on $\Omega\subseteq\mathbb{R}^n$ if for a.e.\, $x\in\Omega$ there is a non-trivial subspace $V_x\subseteq \mathbb{R}^n$ such that $\|F'(x)v\|_2 = \|v\|_2$ for every $v\in V_x$.
\end{definition}
The layers characterized in Section \ref{se:theory} satisfy subspace orthogonality on $\Omega=\mathbb{R}^n$, where $V_x=\mathbb{R}^n$. It is possible to design network layers satisfying this condition on a strictly smaller subspace of $\mathbb{R}^n$ in several ways. In the next lemmas, we discuss two easily implementable constructions.
\begin{lemma}[First construction]\label{lem:first-construction}
Let $F:\mathbb{R}^n\to\mathbb{R}^n$ be Lipschitz and assume that $F'(x)\in\mathcal{O}(n)$ for a.e.\, $x\in\mathbb{R}^n$. 
Let $P\in\mathbb{R}^{n\times n}$ and assume there exists a non-trivial subspace $V\subseteq\mathbb{R}^n$ such that $\|Pv\|_2=\|v\|_2$ for all $v\in V$. Define $G:\mathbb{R}^n\to\mathbb{R}^n$ by $G(x)=PF(x)$. Then $G$ satisfies subspace orthogonality on $\mathbb{R}^n$ with $V_x := (F'(x))^\top V$.
\end{lemma}
\begin{proof}
Fix $x$ such that $F$ is differentiable and $F'(x)\in\mathcal{O}(n)$. Since $G(x)=PF(x)$, we have $G'(x)=PF'(x)$. 
Let $u\in V_x$. Then $u=(F'(x))^\top v$ for some $v\in V$. Therefore, $\|G'(x)u\|_2
=\|PF'(x)(F'(x))^\top v\|_2=\|Pv\|_2=\|v\|_2$.
Moreover, since $F'(x)$ is orthogonal, $\|u\|_2=\|(F'(x))^\top v\|_2=\|v\|_2$. Hence $\|G'(x)u\|_2=\|u\|_2$, which proves subspace orthogonality with subspace $V_x$.
\end{proof}
It is easily verified that if $U\in\mathbb{R}^{n\times m}$ satisfies $U^\top U=I_m$, then the orthogonal projector $P=UU^\top$ satisfies the hypotheses of Lemma~\ref{lem:first-construction} with $V=\mathrm{span}(U)$. In Section \ref{se:experiments} we consider a pool of $k\in \{1,\cdots,n-1\}$ and test
\begin{equation}\label{eq:projection}
P = \begin{bmatrix} I_{n-k} & 0_{n-k,k} \\ 0_{k,n-k} & 0_{k,k} \end{bmatrix}\in\mathbb{R}^{n\times n}.
\end{equation}

\begin{lemma}[Second construction]\label{lem:second-construction}
Let $\alpha\in[0,1)$ and consider the map $F(x)=A^\top \mathrm{LeakyReLU}_{\alpha}(Bx+b)$, with $A,B\in\mathcal{O}(n)$. Define the set
\[
W=\{x\in\mathbb{R}^n:\ (Bx+b)_i\neq 0,\ i=1,\dots,n\},
\]
and, for $x\in W$, the diagonal matrix $D(x)\in\mathbb{R}^{n\times n}$ with
\[
(D(x))_{ii}=\begin{cases}
1 & (Bx+b)_i>0,\\
\alpha & (Bx+b)_i<0,
\end{cases}
\qquad i=1,\dots,n.
\]
Finally, let
\[
\Omega=\Bigl\{x\in\mathbb{R}^n: \exists i\in\{1,\dots,n\}\ \text{s.t. }(Bx+b)_i>0\Bigr\}.
\]
Then $F$ satisfies subspace orthogonality on $\Omega$. More precisely, for every $x\in \Omega\cap W$ the subspace
\[
V_x:=\{v=B^\top u:\ u\in\mathbb{R}^n,\ D(x)u=u\}
\]
is non-trivial and $\|F'(x)v\|_2=\|v\|_2$ for all $v\in V_x$.
\end{lemma}

\begin{proof}
Fix $x\in W$. On $W$, the map $F$ is differentiable and its Jacobian is $F'(x)=A^\top D(x)B$. Let $v\in V_x$. By definition, $v=B^\top u$ for some $u$ such that $D(x)u=u$. Therefore, $\|F'(x)v\|_2=\|A^\top D(x)B(B^\top u)\|_2=\|A^\top D(x)u\|_2 =\|A^\top u\|_2=\|u\|_2=\|B^\top u\|_2=\|v\|_2$, where we used $D(x)u=u$ and that $A,B\in\mathcal{O}(n)$ preserve $\|\cdot\|_2$.
\end{proof}

\begin{remark} The restriction to $\Omega$ ensures that $V_x$ is non-trivial for a.e.\, $x\in\Omega$. Additionally, the set $\mathbb{R}^n\setminus W$ is a finite union of hyperplanes and thus has measure zero.
\end{remark}

\paragraph{Is subspace orthogonality enough for trainability?}
Subspace orthogonality is a layerwise statement and does not control the geometry of the composition $F = F_L\circ\cdots\circ F_1$ of $L$ layers. 
Let $x_1=x$ and $x_{\ell+1}=F_\ell(x_\ell)$. Assume that for every $\ell\in\{1,\dots,L\}$ the layer $F_\ell$ satisfies the subspace orthogonality on $\Omega=\mathbb{R}^n$, i.e., for a.e.\, $x_\ell$ there exists a non-trivial subspace $V^{(\ell)}_{x_\ell}\subseteq \mathbb{R}^n$ such that $\|F_\ell'(x_\ell)v\|_2=\|v\|_2$ for all $v\in V^{(\ell)}_{x_\ell}$. This set of conditions is not sufficient for the product $F'(x)=F_L'(x_L)\cdots F_1'(x_1)$ to preserve the norms on any non-trivial subspace. In other words, contrary to the complete Jacobian orthogonality which is preserved under composition, the passage from layers to the full network is more delicate when considering subspace orthogonality. Indeed, heuristically, to train a network, one needs to ensure the existence of directions spanning a \say{good} subspace in which gradients remain at every depth after being pushed forward by the previous Jacobians. We can identify this space as follows
\begin{equation}\label{eq:Wx}
\begin{split}
W_x \;=\; V^{(1)}_{x_1}&\ \cap\ \big(F_1'(x_1)\big)^{-1}V^{(2)}_{x_2}\\
&\ \cap\ \big(F_2'(x_2)F_1'(x_1)\big)^{-1}V^{(3)}_{x_3}\ \cap\ \cdots\ \\
&\ \cap\ \big(F_{L-1}'(x_{L-1})\cdots F_1'(x_1)\big)^{-1}V^{(L)}_{x_L}.
\end{split}
\end{equation}
By definition, for every $v \in W_x$ it holds $\|F'(x)v\|_2=\|v\|_2$. However, in a typical deep learning setup, the subspaces $V^{(\ell)}_{x_\ell}$ can vary substantially with $\ell$ (e.g., independent random orientations across layers), so that the pulled-back intersection $W_x$ is generally trivial, making layerwise subspace orthogonality unable to prevent vanishing or exploding gradients from appearing. Therefore, to improve network trainability, we need to require a stricter condition: a persistent isometric subspace, ensuring that $W_x$ is non-trivial.
\paragraph{Persistent subspace orthogonality.}
To formalize the general principle expressed in the previous paragraph, let us introduce the following notion.
\begin{definition}[Persistent subspace orthogonality (PSO)]\label{def:pso}
Let $L\in\mathbb{N}$, and consider the map $F=F_L\circ \cdots \circ F_1$ with $F_\ell:\mathbb{R}^n\to\mathbb{R}^n$ Lipschitz continuous, $\ell=1,\cdots,L$. Let $\Omega\subseteq\mathbb{R}^n$, and, for every $x\in\Omega$, we introduce the notation $x_{\ell+1}=F_\ell(x_\ell)$, $x_1=x\in\Omega$. We say that $F$ satisfies the persistent subspace orthogonality (PSO) over $\Omega$ if:
\begin{enumerate}
    \item for $\ell=1,\cdots,L$ and a.e.\, $x\in\Omega$ there is a non-trivial subspace $V_{x_\ell}^{(\ell)}\subseteq\mathbb{R}^n$ with $\|F_\ell'(x_\ell)v\|_2=\|v\|_2$ for every $v\in V_{x_\ell}^{(\ell)}$,
    \item $W_x$ defined in \eqref{eq:Wx} is non-trivial for a.e.\, $x\in\Omega$.
\end{enumerate}
\end{definition}

\begin{figure*}[ht!]
\centering
\includegraphics[width=0.9\textwidth]{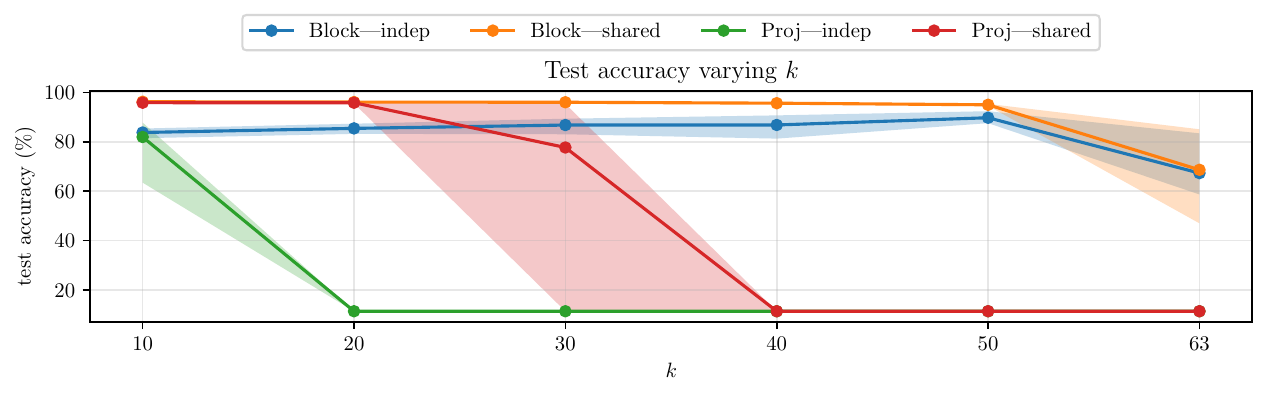}
\caption{\textbf{Sensitivity to the dimension $k$ of the non-isometric subspace.} MNIST results for projection and block variants.}
\label{fig:MNIST_varyK}
\end{figure*}

PSO mitigates vanishing gradients because, if $\Omega$ contains (part of) the dataset, along the subspace $W_x$, backpropagation preserves the norm of gradient components across depth at the beginning of training. Assuming $\dim(W_x)$ is sufficiently large, typical gradients admit non-trivial projections onto $W_x$, so a non-negligible fraction of the backpropagated signal is preserved. Nonetheless, we show in the experiments (see Figure \ref{fig:MNIST_varyK}) that $\dim(W_x)=1$ can be sufficient. 
\begin{remark}
The PSO condition does not prescribe the Jacobian behavior in the orthogonal complements to the isometric subspaces $W_x$. Still, all the initializations presented in this section yield $1$-Lipschitz maps whose Jacobian spectral norm is uniformly bounded by $1$. Therefore, within the PSO framework, we can mitigate both vanishing and exploding gradients.
\end{remark}

\paragraph{Two strategies to ensure a persistent subspace} We now provide two methodologies to initialize neural networks satisfying the PSO condition. The first strategy shows that whenever the explicit form of the input-to-output Jacobian of the considered network can be directly constrained, the PSO condition reduces to the subspace orthogonality of a single map. The second approach provides, instead, a more general procedure that enforces subspace orthogonality layerwise.

\makeatletter
\def\@currentlabel{Approach 1}\label{app1}
\makeatother
\textit{Approach 1.} Consider a network with $L$ layers of the form
\[
F_{\theta_\ell}(x) = A^\top_\ell \mathrm{LeakyReLU}_{\alpha}(B_\ell x + b_\ell),\,\,\theta_\ell = (A_\ell,B_\ell,b_\ell),
\]
where $\ell=1,\cdots,L$, $\alpha\in [0,1)$. Fixing $\bar{\theta}_\ell=(Q,Q,0_n)$, $Q\in\mathcal{O}(n)$, for every $\ell=1,...,L$, one can verify that
\begin{equation}\label{eq:idempotent}
F_{\bar{\theta}}(x)=F_{\bar{\theta}_L}\circ \cdots \circ F_{\bar{\theta}_1}(x)  = Q^\top \sigma(Qx;\alpha^L).
\end{equation}
Lemma \ref{lem:second-construction} shows that the maps as \eqref{eq:idempotent} are subspace orthogonal. Given that the breaklines of the piecewise affine function $F_{\bar{\theta}}$ are the same as the individual layers, $F_\theta$ can be initialized to satisfy PSO, i.e., as $F_{\bar{\theta}}$ for a $Q\in\mathcal{O}(n)$ drawn at random. Table \ref{tab:ffMNIST} validates the effectiveness of this strategy for four values of $\alpha\in\{0,0.01,0.1,0.2\}$. 

\makeatletter
\def\@currentlabel{Approach 2}\label{app2}
\makeatother
\textit{Approach 2.} The second strategy that we propose enforces, at initialization, a prescribed and shared isometric subspace across layers. Fix $k\in\{0,\dots,n-1\}$ and let $U\in\mathbb{R}^{n\times n-k}$ have orthonormal columns, $U^\top U=I_{n-k}$. Denote $V=\operatorname{span}(U)$. We can construct layers of the form
\begin{equation}\label{eq:splitOrthogonal}
F_{\theta_\ell}(x)=U\,G_{\theta_{\ell}}(U^\top x),\,\,G_{\theta_{\ell}}:\mathbb{R}^{n-k}\to\mathbb{R}^{n-k},
\end{equation}
for $\ell=1,\dots,L$. Whenever $G_{\theta_{\ell}}$ is differentiable, the chain rule gives $F'_{\theta_\ell}(x)U = U G_{\theta_{\ell}}'(U^\top x)$. Assume now that $G'_{\theta_{\ell}}(y)\in\mathcal{O}(n-k)$ for a.e.\, $y\in\mathbb{R}^{n-k}$. Then, for a.e.\, $x\in\mathbb{R}^n$ and an arbitrary $v\in\mathbb{R}^n$, we have $UU^\top v\in V$ and
\begin{align*}
&\|F'_{\theta_\ell}(x)UU^\top v\|_2
=\|UG_{\theta_{\ell}}'(U^\top x)U^\top v\|_2 =\|G_{\theta_{\ell}}'(U^\top x)U^\top v\|_2=\|U^\top v\|_2=\|UU^\top v\|_2,
\end{align*}
where we used the assumptions $U^\top U=I_{n-k}$ and $G_{\theta_{\ell}}'(U^\top x)\in\mathcal{O}(n-k)$. This reasoning implies that each layer is subspace orthogonal on the fixed subspace $V$ and $F_\theta = F_{\theta_L}\circ \cdots \circ F_{\theta_1}$ satisfies the PSO condition.

In our experiments, we consider the case
\[
U=\begin{bmatrix} I_{n-k} \\ 0_{k\times(n-k)}\end{bmatrix},\,\,
G_{\theta_{\ell}}(y) = A^\top_\ell |B_\ell y+b_\ell|\in\mathbb{R}^{n-k},
\]
for a choice of $A_\ell,B_\ell\in\mathcal{O}(n-k)$ and $b_\ell\in\mathbb{R}^m$. We demonstrate that the proposed models train even with $k=n-1$, i.e., with a one-dimensional isometric subspace (Figure \ref{fig:MNIST_varyK}). We remark that this modeling strategy can be seen as a block structured initialization of a layer on $\mathbb{R}^n$, which is how we define our layers:
\[
x \mapsto \begin{bmatrix} B_\ell & 0_{n-k, k}\\ 0_{k, n-k} & 0_{k, k} \end{bmatrix}\left |\begin{bmatrix} A_\ell & 0_{n-k, k}\\ 0_{k, n-k} & 0_{k, k} \end{bmatrix} x + \begin{bmatrix} b_\ell \\ 0_k \end{bmatrix} \right |.
\]
\begin{remark}[Input and output dimensions]
We presented characterizations of subspace orthogonality and PSO for maps that preserve the input dimension. This choice is made in continuity with Section \ref{se:theory} on exact Jacobian orthogonality. However, subspace orthogonality has no intrinsic limitation on the input and output dimensions.
\end{remark}
\begin{remark}
We experimentally show that PSO need not be maintained during training: a PSO-satisfying initialization already enables training at depth. We therefore only need to constrain the weights at initialization to enter a PSO regime, then optimize all parameters freely, in line with how dynamical isometry is generally intended.
\end{remark}

\begin{table*}[ht!]
  \centering
  \caption{\textbf{Sensitivity to network depth $L$ for feedforward models.} MNIST results for feedforward models (numbers indicate percentages). The parameter $\alpha$ represents the negative slope in the activation $\mathrm{LeakyReLU}_{\alpha}$, i.e., $\alpha=0$ coincides with $\mathrm{ReLU}$. The numbers after $\pm$ represent the maximum absolute discrepancy from the mean across the five seeds.}
  \label{tab:ffMNIST}
  \vspace{10pt}
  \resizebox{\textwidth}{!}{%
    \setlength{\tabcolsep}{4pt} 
      \begin{tabular}{c|cccccc}
        \toprule
        $L$ & \shortstack{\textbf{Kaiming}\\($\alpha=0$)} & \shortstack{\textbf{Orth Indep}\\($\alpha=0$)} & \shortstack{\textbf{Orth Shared}\\($\alpha=0$)} & {\shortstack{\textbf{Orth Shared}\\($\alpha$=0.01)}} & {\shortstack{\textbf{Orth Shared}\\($\alpha$=0.1)}} & \shortstack{\textbf{Orth Shared}\\($\alpha$=0.2)} \\
        \midrule
        \textbf{5}   & 96.35 $\pm$ 0.30 & 96.37 $\pm$ 0.30 & 96.72 $\pm$ 0.29 & 96.73 $\pm$ 0.25 & 96.65 $\pm$ 0.26 & 96.54 $\pm$ 0.22 \\
        \textbf{20}  & 90.06 $\pm$ 1.62 & 94.41 $\pm$ 0.33 & 97.09 $\pm$ 0.08 & 97.12 $\pm$ 0.10 & 97.09 $\pm$ 0.15 & {97.17 $\pm$ 0.17} \\
        \textbf{50}  & 12.00 $\pm$ 0.96 & 91.49 $\pm$ 2.52 & 97.22 $\pm$ 0.16 & 97.22 $\pm$ 0.16 & 97.19 $\pm$ 0.08 & {97.27 $\pm$ 0.15} \\
        \textbf{100} & 10.17 $\pm$ 0.46 & 11.35 $\pm$ 0.00 & 97.19 $\pm$ 0.10 & {97.26 $\pm$ 0.07} & 97.18 $\pm$ 0.13 & 97.22 $\pm$ 0.11 \\
        \textbf{200} & 10.01 $\pm$ 1.30 & 11.35 $\pm$ 0.00 & 97.20 $\pm$ 0.10 & 97.11 $\pm$ 0.18 & {97.21 $\pm$ 0.10} & 97.18 $\pm$ 0.14 \\
        \bottomrule
      \end{tabular}
      }
\end{table*}

\section{Numerical experiments}\label{se:experiments}
In our experiments, we focus on classification tasks for three datasets: MNIST, Fashion-MNIST, and CIFAR-10. The results for MNIST are in this section. The two other datasets are considered in the three ablation studies in Section \ref{app:ablations}. There, we show that the initialization strategy for the first and last linear layers is not essential, demonstrate the practicality of the activations $\mathrm{ReLU}_3$ and $\sigma_3$ introduced in \eqref{eq:stair} and \eqref{eq:generalAbs}, and assess the network trainability across different learning rates. For CIFAR-10 and Fashion-MNIST, we consider convolutional neural networks. See Appendix \ref{app:experiments} for more details on the experimental setup.

\paragraph{Sensitivity to the network depth $L$.}
We consider different network depths $L\in\{5,20,50,100,200\}$ and compare feedforward versus residual blocks of layers under multiple initialization and activation choices. As activations, we consider $\mathrm{LeakyReLU}_{\alpha}$ with $\alpha\in\{0,0.01,0.1,0.2\}$. 

Each network maps $28\times 28$ images to a width-$d$ hidden space with a linear input layer ($784 \to d$), applies $L$ identical-width blocks, and ends with a linear $10$-way classifier. Each block computes
\[
z = Bx + b,\quad h = \mathrm{LeakyReLU}_{\alpha}(z),\quad u = A^\top h,
\]
and outputs $u$ (feedforward) or $x+u$ (residual). We consider the models of types \ref{eq:type_A} and \ref{eq:type_B}, as well as a gradient residual variant that fixes the residual map to use $ A =- 2B$ while keeping standard residual addition, as in \eqref{eq:ReLUk}. The feedforward models with shared orthogonal initialization all satisfy the PSO condition, see \ref{app1}. The residual models \textit{Orth-2} and \textit{Negative Gradient} have orthogonal Jacobians a.e.\, at initialization, see Section \ref{se:theory}.


\paragraph{Testing the dimension of the orthogonality subspace.}
We fix depth at $L=100$, width $d=64$, and use the absolute value activation with orthogonal initialization in feedforward networks, i.e., layers of the form $F_{\theta_\ell}(x)=A_\ell^\top|B_\ell x + b_\ell|$. Two structural constraints are studied: a learnable projection $P$ after each block as in Lemma \ref{lem:first-construction}, initialized to preserve the first $d-k$ coordinates, and a block-structured orthogonal initialization in which only the top-left $(d-k)\times(d-k)$ sub-block is orthogonal, and all other entries are zero, as in \ref{app2}. We consider different values for $k\in\{10,20,30,40,50,63\}$. We remark that the models with block initialization all satisfy the PSO condition as described in \ref{app2}. The projected models have only subspace orthogonal layers (see Lemma \ref{lem:first-construction}), but do not guarantee satisfaction of the PSO condition.

\paragraph{Results.} We start from the experiment with a \textit{varying depth} $L$. We report in Tables~\ref{tab:ffMNIST} and \ref{tab:resnetMNIST} the mean test accuracies together with the confidence interval provided by the maximum absolute deviation from the mean. The results demonstrate the improved trainability of models that satisfy the PSO condition. It is also clear that only having layers which are subspace orthogonal is not enough, as can be seen from the poor results at $L\in\{100,200\}$ for the feedforward model with independent orthogonal weights, see Table~\ref{tab:ffMNIST}.

Figure~\ref{fig:MNIST_varyK} collects the results on the sensitivity to the dimension of the isometric subspace. The projected models, regardless of how the weights are initialized, are very unstable and sensitive to the dimension of the isometric subspace. This follows from the inability to guarantee a persistent subspace across the layers. The block-structured layers are very robust to higher values of $k$ and reliably train also with a one-dimensional persistent isometric subspace ($k=63$).

\begin{table*}[ht!]
  \centering
  \caption{\textbf{Sensitivity to network depth $L$ for residual models.} MNIST results for ResNet models (numbers indicate percentages). The parameter $\alpha$ represents the negative slope in the activation $\mathrm{LeakyReLU}_{\alpha}$, i.e., $\alpha=0$ coincides with $\mathrm{ReLU}$. \textit{Orth-2} refers to the initialization with $B$ random orthogonal, and $A=-2B$. \textit{Negative Gradient} refers to networks having layers $x\mapsto x-2B^\top\mathrm{LeakyReLU}_\alpha(Bx+b)$ where $B$ is a random orthogonal matrix. We write $\mathrm{NaN}$ if the loss diverges with at least one of the seeds. The numbers after $\pm$ represent the maximum absolute discrepancy from the mean across the five seeds.}
  \label{tab:resnetMNIST}
  \vspace{10pt}
  \setlength{\tabcolsep}{10pt}
  \begin{tabular}{c|cccc}
    \toprule
    $L$ & \shortstack{\textbf{Kaiming}\\($\alpha=0$)} & \shortstack{\textbf{Orth Indep}\\($\alpha=0$)} & \shortstack{\textbf{Orth-2}\\($\alpha=0$)} & \shortstack{\textbf{Negative Gradient}\\($\alpha=0$)} \\
    \midrule
    \textbf{5}   & 96.26 $\pm$ 0.10 & {97.01 $\pm$ 0.16} & 96.97 $\pm$ 0.17 & 96.83 $\pm$ 0.20 \\
    \textbf{20}  & 75.55 $\pm$ 4.17 & 95.35 $\pm$ 0.25 & 96.72 $\pm$ 0.26 & 96.88 $\pm$ 0.32 \\
    \textbf{50}  & 19.16 $\pm$ 3.87 & 77.00 $\pm$ 3.25 & 96.23 $\pm$ 0.34 & 96.30 $\pm$ 0.21 \\
    \textbf{100} & 10.36 $\pm$ 1.34 & 24.46 $\pm$ 9.43 & 93.45 $\pm$ 0.48 & 92.55 $\pm$ 0.68 \\
    \textbf{200} & NaN & 11.29 $\pm$ 0.64 & 83.63 $\pm$ 1.79 & 83.10 $\pm$ 3.66 \\
    \bottomrule
  \end{tabular}
\end{table*}

\section{Ablation studies}\label{app:ablations}
We identify three components that may affect the learning process and assess their impact on the trainability of the proposed models. These ablation studies demonstrate the crucial role of PSO relative to other factors. Further information on the setup of the experiments is detailed in Appendix \ref{app:experiments}.

\subsection{Role of the initialization of the first and last affine layers}\label{app:ablation_first_and_last}
\begin{figure}[h!]
    \centering
    \includegraphics[width=0.9\linewidth]{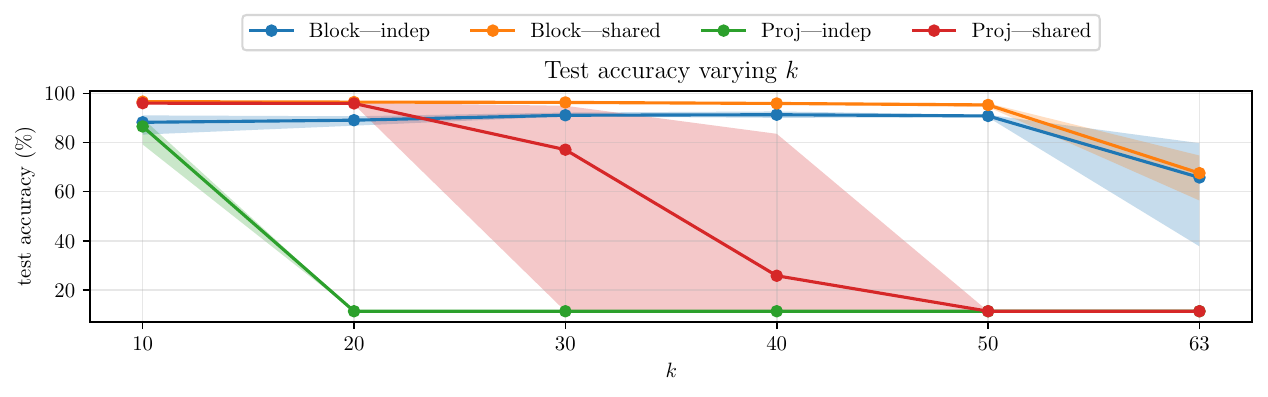}
    \caption{\textbf{Ablation over the initialization of the first and last affine layers.} MNIST results for projection and block variants for non-orthogonal dimensionality reduction layers.}
    \label{fig:MNIST_varyK_non_orth}
\end{figure}

In this experiment, we work with the MNIST dataset. In the $k$-varying experiment in Figure \ref{fig:MNIST_varyK}, we considered networks relying on two dimensionality reduction layers (at the beginning and at the end of the networks). The first takes vectorized images from $\mathbb{R}^{784}$ to $\mathbb{R}^{64}$, and the last goes from $\mathbb{R}^{64}$ to the logits in $\mathbb{R}^{10}$. For those experiments, we initialized such matrices as Stiefel matrices, i.e., as matrices defining isometries of the arrival space. This allows the full input-to-output map to be isometric on the arrival space for the block-stuctured architectures given that the hidden layers have orthogonal Jacobians. We repeat here the same experiments but initializing the entries of such matrices from the uniform distribution over $[-\sqrt{\lambda},\sqrt{\lambda}]$, $\lambda=1/64$, which is PyTorch's default initialization. Comparing Figures \ref{fig:MNIST_varyK} and \ref{fig:MNIST_varyK_non_orth}, we see that this choice appears to play a marginal role for the trainability of very deep networks, supporting the claim that the model's trainability is mostly determined by the properties of the hidden layers and therefore that having them satisfy the PSO property is crucial.

\subsection{Sensitivity to the activation function}\label{app:ablation_act_fn}
In this experiment, we work with the Fashion-MNIST dataset. In previous experiments, we always employed simple nonlinearities such as $\mathrm{ReLU}$ or the absolute value function. However, the results provided by Theorem \ref{thm:char_RtoR_fields} open up to a much wider family of activation functions. To test the practicality of some of them, we consider
\begin{enumerate}[label=\arabic*)]
    \item for case \ref{ci}, $x \mapsto A^\top \rho(Bx + b)$ with $\rho\in\{\sigma_1, \sigma_3\}$;
    \item for case \ref{cii}, $x \mapsto x - B^\top \rho(Bx + b)$ with $\rho \in \{2\mathrm{ReLU}, 2\mathrm{ReLU}_3\}$;
\end{enumerate}
where the action of the kernels is realized through convolutions. We recall that $\mathrm{ReLU}_3$ is defined in \eqref{eq:stair}, and $\sigma_3$ in \eqref{eq:generalAbs}. The results are reported in Table \ref{tab:conv_pw}. In agreement with our theory, different activation functions with suitably constrained slopes do not undermine the training of the underlying architectures since they still allow for exact Jacobian orthogonality.

\begin{table}[h]
  \centering
  \caption{\textbf{Ablation over different activation functions.} The table reports the classification accuracy on the validation set of Fashion-MNIST. We test models with $L$ layers, $L\in\{10,50,200\}$.}
  \label{tab:conv_pw}
  \vspace{10pt}
  \setlength{\tabcolsep}{10pt}     
  \begin{tabular}{c|cccc}
    \toprule
    $L$ & \textbf{FF-}$\mathbf{\sigma_1}$ & \textbf{FF-}$\mathbf{\sigma_3}$ & \textbf{ResNet‐}$\mathbf{\mathrm{ReLU}}$ & \textbf{ResNet‐}$\mathbf{\mathrm{ReLU}_3}$\\
    \midrule
    \textbf{10} & 86.8\% & 89.3\% & 87.2\% & 89.7\% \\
    \textbf{50} & 87.1\% & 83.7\% & 88.8\% & 87.8\% \\
    \textbf{200} & 86.0\% & 84.3\% & 88.2\% & 89.1\% \\
    \bottomrule
  \end{tabular}
\end{table}

\subsection{Comparison across learning rates}\label{app:ablation_lr}
\begin{figure}[h!]
    \centering
    \includegraphics[width=0.9\linewidth]{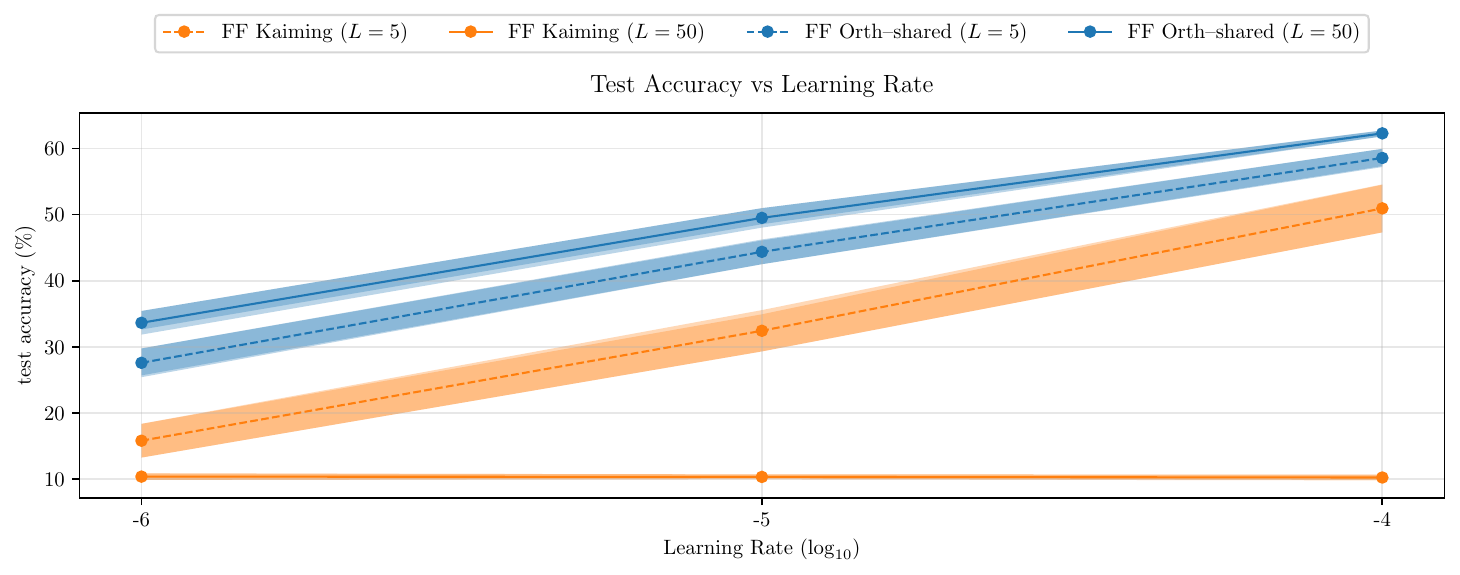}
    \caption{\textbf{Ablation over different learning rates.} Comparison of the test accuracies of four feedforward networks with weights initialized as Kaiming normal and as shared orthogonal weights. We consider two depths, $L \in \{5, 50\}$, and fixed learning rates in $\{10^{-6},10^{-5},10^{-4}\}$. All the models are trained for $20$ epochs.}
    \label{fig:comparison_KN_vs_Orth_shared}
\end{figure}

In this experiment, we work with the CIFAR-10 dataset. An important aspect of network trainability is the choice of the learning rate. We consider two convolutional feedforward models of the form \eqref{eq:type_A}, one with filters initialized as Kaiming normal, and the other as proposed in \ref{app1}, i.e., with shared orthogonal linear maps realized via convolution. We fix the number of epochs to 20. Then, we compare the test accuracies at depth $L \in \{5, 50\}$ for different learning rates. The results shown in Figure \ref{fig:comparison_KN_vs_Orth_shared} suggest a consistent performance increase with PSO network initializations. More explicitly, independent of the learning rate, models initialized with Kaiming normal distributions are more challenging to train than those initialized with PSO, which start training effectively across all regimes.

\section{Discussion and future work}\label{se:discussion}
This work identifies a necessary and concrete mechanism supporting the trainability of very deep networks: the existence of a non-trivial subspace on which layer Jacobians act isometrically persistently across depth. We provide a characterization of a class of layers whose Jacobians are orthogonal almost everywhere, and we introduce simple and implementable constructions that enforce persistent subspace orthogonality (PSO) at initialization. Empirically, we find that preserving PSO throughout training is unnecessary in the settings considered: initializing in a PSO regime already stabilizes optimization, consistent with the dynamical isometry principle. We validate these effects across a range of depths and architectures, including feedforward, residual, and convolutional models.

Our theoretical characterizations are currently sharpest for layers that are globally (almost everywhere) isometric, i.e.\ with orthogonal Jacobians. An important extension is to develop analogous results for the PSO setting, where isometry holds only on a prescribed subspace.

A second direction is to transfer these initialization principles to more structured architectures such as Transformers and graph neural networks, where stability of deep compositions and the geometry of Jacobians are also central. Designing PSO-type initializations compatible with attention and message passing is a natural next step.

Finally, \ref{app1} highlights that, for weight-tied constructions with suitable nonlinearities, repeated composition admits closed-form simplifications (e.g.\ an effective depth parameter). This suggests a mechanism by which training can progressively \say{activate} depth, and motivates the study of depth-adaptation schemes and their interactions with optimization and generalization.


\bibliographystyle{plain}
\bibliography{bib}

\newpage

\appendix

\onecolumn
\section{Proof of Theorem \ref{thm:char_RtoR_fields}}\label{app:proofThm1}
In this appendix, we use the notation 
\[
\begin{aligned}
\mathcal{D}(\alpha_1,\alpha_2):=\{D=\mathrm{diag}(d_1,\cdots,d_n)\in\mathbb{R}^{n\times n}:\,d_1,\cdots,d_n\in\{\alpha_1,\alpha_2\}\},
\end{aligned}
\]
with $\alpha_1,\alpha_2\in\mathbb{R}$, to denote the set of diagonal matrices with entries in a set of two elements $\{\alpha_1,\alpha_2\}$.

Consider an arbitrary partition $\{\Omega_1,\cdots,\Omega_N\}$ in $\mathcal{P}(\mathbb{R}^n)$. To simplify the notation and keep track of the slopes, if $\sigma_i\in\mathcal{L}(\alpha_i,\beta_i)$, we write $\sigma_i(x)=\sigma(x;\alpha_i,\beta_i)$.  When considering the derivative $\sigma_i'(x)$, we use the notation $\sigma'(x;\alpha_i,\beta_i)$. Furthermore, we denote with $F_{\Omega_i}'(x)\in\mathbb{R}^{n\times n}$, $i=1,\cdots,N$, the Jacobian matrix of the restriction $F_{\Omega_i}:\Omega_i\to\mathbb{R}^n$ of $F$ to $\Omega_i$. 
\paragraph{Verification that \ref{ci} and \ref{cii} have orthogonal Jacobians:} 
\textbf{Case \ref{ci}:} Let $x\in\Omega_i$. We have
\[
F_{\Omega_i}(x)=d_i A^\top \sigma\left(Bx + b\,;-\frac{1}{d_i},\frac{1}{d_i}\right) + c_i
\]
and hence
\[
F_{\Omega_i}'(x) = d_iA^\top \mathrm{diag}\left(\sigma'\left(Bx+b\,;-\frac{1}{d_i},\frac{1}{d_i}\right)\right)B=: d_iA^\top D_i(x) B.
\]
Because of the properties of $\sigma$, for a.e.\, $x\in\Omega_i$, $D_i(x)\in \mathcal{D}(-1/d_i,1/d_i)$. We conclude
\[
F_{\Omega_i}'(x)^\top F_{\Omega_i}'(x) = d_i^2 B^\top (D_i(x))^2 B= d_i^2 B^\top (I_n / d_i^2)B = I_n.
\]

\textbf{Case \ref{cii}:} We have
\[
F_{\Omega_i}(x)=\ell_ix + d_i B^\top \sigma\left(Bx + b\,;\frac{1-\ell_i}{d_i},-\frac{1+\ell_i}{d_i}\right) + c_i
\]
and hence
\[
\begin{split}
F_{\Omega_i}'(x)^\top F_{\Omega_i}'(x)& = (\ell_iI_n + d_iB^\top D_i(x)B)^\top (\ell_iI_n + d_iB^\top D_i(x)B) \\
&= \ell_i^2I_n + d^2_i B^\top (D_i(x))^2 B + 2\ell_id_iB^\top D_i(x)B\\
&=B^\top \left(\ell_i^2I_n + d^2_i (D_i(x))^2 + 2 \ell_id_iD_i(x)\right)B\\
&= B^\top \left(\ell_i I_n + d_i D_i(x)\right)^2B = B^\top \widetilde{D}_i(x)^2 B\\
& = B^\top B = I_n
\end{split}
\]

where $D_i(x)\in \mathcal{D}((1-\ell_i)/d_i, -(1+\ell_i)/d_i)$ and $\widetilde{D}_i(x)\in \mathcal{D}(\pm 1)$.

\paragraph{Verification that \ref{ci} and \ref{cii} are the only options:}

Let us consider the function $F$ defined as in \eqref{eq:focusMap} and restrict to one set $\Omega_i$ of an arbitrary admissible partition $\{\Omega_1,\cdots,\Omega_N\}$ so that it takes the form
\begin{equation}\label{eq:ithPiece}
F_{\Omega_i}(x) = \ell_i x + c_i + d_i A^\top \sigma_i(Bx+b).
\end{equation}
Since $\sigma_i:\mathbb{R}\to\mathbb{R}$ is Lipschitz continuous and piecewise $C^1$, so is $F_{\Omega_i}$. Thus, Lemma \ref{lemma:char_orth_field} allows us to infer that $\sigma_i$ is piecewise affine. 

The Jacobian matrix of \eqref{eq:ithPiece} is
\[
\begin{aligned}
F_{\Omega_i}'(x) & = \ell_i I_n + d_i A^\top \mathrm{diag}(\sigma'_i(Bx + b))B\\
& = \ell_i I_n + d_iA^\top D_i(x) B,
\end{aligned}
\]
where we set $D_i(x) = \mathrm{diag}(\sigma_i'(Bx+b))$. We remark that, despite $\sigma_i'$ has $B$ as an input, $B$ is invertible and hence for any $y\in\mathbb{R}^n$, there is $x\in\mathbb{R}^n$ with $Bx+b=y$, and hence we can cover any value of $\sigma'(\mathbb{R})$ up to properly choosing $x$. Such a point $x$ can be assumed to be in $\Omega_i$, since this result must hold for any partition. This allows us to assume $D_i(x)$ is independent of $B$, and, clearly, also independent of $A$. The conditions $F_{\Omega_i}'(x)^\top F_{\Omega_i}'(x) = F_{\Omega_i}'(x)F_{\Omega_i}'(x)^\top = I_n$ write
\[
\begin{aligned}
&(\ell^2_i - 1) I_n + d_i^2A^\top (D_i(x))^2 A + \ell_i d_i(A^\top D_i(x)B+ B^\top D_i(x)A) = 0,\\
&(\ell^2_i - 1) I_n + d_i^2B^\top (D_i(x))^2 B + \ell_i d_i (A^\top D_i(x)B+ B^\top D_i(x)A) = 0.
\end{aligned}
\]
Subtracting the two conditions above, we get the necessary condition for orthogonality
\[
A^\top (D_i(x))^2 A = B^\top (D_i(x))^2 B
\]
or, equivalently,
\begin{equation}\label{eq:commuting}
(AB^\top) (D_i(x))^2 = (D_i(x))^2 (AB^\top).
\end{equation}
We remark that our result must be true for an arbitrary partition $\{\Omega_1,\cdots,\Omega_N\}$ and hence for an arbitrary open and connected set $\Omega_i\subseteq\mathbb{R}^n$ as well. Since $A, B \in \mathcal{O}(n)$ are arbitrary orthogonal matrices but they are shared among the sets $\Omega_1,\cdots,\Omega_N$, we must either have $AB^\top = P$, with $P \in \mathcal{D}(\pm1)$, or $(D_i(x))^2 = b_i I_n$, for every $i=1,\cdots,N$, for a.e.\, $x\in\mathbb{R}^n$ a suitable choice of $N$ non-negative scalars $b_1,\cdots,b_N\in\R$. In fact, if there exists $i\in\{1,\cdots,N\}$ with $(\sigma_i'(\mathbb{R}))^2$ which is not a singlet, since $\Omega_i$ is arbitrary, we can assume that for every $r\neq s$, $r,s\in\{1,\cdots,n\}$, there exists $\Omega_i\subseteq\mathbb{R}^n$ admitting a positive-measure subregion $\mathcal{R}\subseteq\Omega_i$ with $(D_i(x)^2)_{rr}\neq (D_i(x)^2)_{ss}$ for every $x\in\mathcal{R}$, hence leading to the constraint that $AB^\top$ must be diagonal (and orthogonal).

We now consider each case separately and show that they lead to the two forms in the theorem.

\paragraph{Case $A=PB$.} We now have $F_{\Omega_i}'(x)^\top = F_{\Omega_i}'(x)$ and the orthogonality conditions reduce to
\begin{equation}\label{eq:identity}
\left(\ell_i I_n + d_i PD_i(x)\right)^2 
= I_n.
\end{equation}

\textit{Assume $\ell_i\neq 0$.} Then either $D_i(x)=0$ for every $x\in\mathbb{R}^n$ and for every $i=1,\cdots,n$, in which case $D_i(x)^2 = b_i I_n$ (and we treat it below), or there is an $i\in\{1,\cdots,N\}$ and a positive-measure subregion of $\mathbb{R}^n$ where $D_i(x)\neq 0$. When the second case verifies, either $P=I_n$ or $P=-I_n$. In fact, if $P$ is any other matrix in $\mathcal{D}(\pm 1)$, since the partition is arbitrary, we can assume that there exists a pair of indices $r\neq s$, $r,s\in\{1,\cdots,n\}$, such that $P_{rr}=1=-P_{ss}$ and $(D_i(x))_{rr}=(D_i(x))_{ss}\neq 0$ for any $x\in\mathcal{R}$, $\mathcal{R}\subseteq\Omega_i$ a positive-measure subregion. This leads to a contradiction since it must be true that
\[
\begin{aligned}
& \ell_i + d_i(D_i(x))_{rr}\in\{\pm 1\},\\
& \ell_i - d_i(D_i(x))_{ss}=\ell_i - d_i(D_i(x))_{rr}\in\{\pm 1\}.
\end{aligned}
\]
A necessary condition for this to hold is that $\ell_i=0$, which is not allowed in this configuration. 

If $P=I_n$, \eqref{eq:identity} together with $\ell_i\neq 0$ imply that $D_i(x)\in \mathcal{D}((\pm 1 - \ell_i)/d_i)$, which fully characterizes case \ref{cii}. If $P=-I_n$, we recover $D_i(x)\in\mathcal{D}((\pm 1+\ell_i)/d_i)$. However, this second situation is equivalent to the former, since we can call $d_i' = -d_i$ and recover the same condition $D_i(x)\in\mathcal{D}((\pm 1-\ell_i)/d_i')$, and the layer takes the desired form 
\[
F_{\Omega_i}(x) = \ell_ix+c_i - d_i B^\top \sigma_i(Bx+b)= \ell_ix+c_i+d_i'B^\top \sigma_i(Bx+b)
\]
where $\mathrm{diag}(\sigma_i'(Bx+b))\in\mathcal{D}((\pm 1-\ell_i)/d_i')$, i.e., case \ref{cii}.

\textit{Assume, instead, $\ell_i=0$.} In this situation, it is enough to have $d_iD_i(x)\in\mathcal{D}(\pm 1)$, and hence the condition $D_i(x)\in\mathcal{D}(\pm 1/d_i)$, as in case \ref{ci}, follows. Here, the matrix $P$ can be kept generic in $\mathcal{D}(\pm 1)$, and we recover a particular instance of case \ref{ci}, since instead of having $A$ independent from $B$, we have it equal to $PB$.

\paragraph{Case $(D_i(x))^2=b_i I_n$.} Here, the orthogonality conditions reduce to the single equation
\begin{equation}\label{eq:simplifiedEq}
(\ell^2_i - 1) I_n + b_id_i^2I_n + \ell_i d_i(A^\top D_i(x)B + B^\top D_i(x)A) = 0.
\end{equation}

\textit{Assume $\ell_i=0$.} We recover $-1+b_id_i^2=0$ and hence $b_i = 1/d_i^2$. This allows us to derive the desired condition, i.e., $D_i(x)\in\mathcal{D}(\pm 1/d_i)$, fully characterizing case \ref{ci}.

\textit{Assume, instead, $\ell_i\neq 0$.} We have
\begin{equation*}
A^\top D_i(x)B + B^\top D_i(x)A = k_iI_n
\end{equation*}
where we set $k_i = -\frac{\ell_i^2 - 1 + b_id_i^2}{\ell_i d_i}$. Then, there exists a skew-symmetric matrix valued function $N_{i}(\cdot;A,B):\R^n\to\R^{n\times n}$, $N_i(x;A,B)^\top =-N_i(x;A,B)$, possibly depending on $A$ and $B$, such that
\begin{equation}\label{eq:secondToLast}
A^\top D_i(x)B = \frac{k_i}{2}I_n + N_i(x;A,B).
\end{equation}
\eqref{eq:secondToLast} tells us that $N_i$ depends continuously on the parameters $A$ and $B$, since they appear linearly in the left hand side and it must be true that $N_i(x;A,B)=A^\top D_i(x)B-k_i/2 I_n$. This condition is equivalent to
\begin{equation}\label{eq:lastCase}
D_i(x) = \frac{k_i}{2}AB^\top + AN_i(x;A,B)B^\top =: \mathscr{S}(x, A, B).
\end{equation}
We notice that, since the left-hand side of \eqref{eq:lastCase} is independent of $A$, the same should hold for $\mathscr{S}(x, \cdot, B): \mathcal{O}(n) \to \mathbb{R}^{n \times n}$. In particular, since $\mathcal{O}(n)$ has two connected components and $\mathscr{S}(x, A, B)$ is continuous in $A$, this means that $\mathscr{S}(x, \cdot, B)$ is constant on each of these. Let us denote $SO(n)$ the subset of $\mathcal{O}(n)$ of those matrices with positive determinants. Thus, without loss of generality we can consider $B \in SO(n)$ and set $A=B$, which gives
\[
D_i(x) = \frac{k_i}{2}I_n + BN_i(x;B,B)B^\top.
\]
since $D_i(x)-k_i/2 I_n$ is symmetric whereas $BN_i(x;B,B)B^\top$ is skew-symmetric and similar to $N_i(x;B,B)$, we conclude $N_i(x;B,B)=0$ for any $x\in\mathbb{R}^n$ and $B\in\mathcal{O}(n)$. Thus 
\[
D_i(x) = \frac{k_i}{2}I_n
\]
which is independent of $A$ (and $B$) as wanted. Analogously, given $A = -B \in \mathcal{O}(n) \setminus SO(n)$, we conclude
\[
D_i(x) = -\frac{k_i}{2}I_n.
\]
In both cases, we obtain $b_i = k_i^2/4$.

On the other hand, \eqref{eq:secondToLast} also implies
\[
\frac{k_i^2}{4}I_n = b_iI_n = B^\top D_i(x)^2B= (A^\top D_i(x)B)^\top(A^\top D_i(x)B)= \frac{k_i^2}{4}I - N_i(x;A,B)^2
\]
and hence $N_i(x;A,B)^2=0$ for every $A,B\in\mathcal{O}(n)$. Since 
\[
N_i(x;A,B)^2=-N_i(x;A,B)^\top N_i(x;A,B)
\]
is symmetric negative semi-definite, and it vanishes, we must have $N_i(x;A,B)=0$ for every $x\in\mathbb{R}^n$ and $A,B\in\mathcal{O}(n)$ for which \eqref{eq:secondToLast} is true.
Substituting back in \eqref{eq:lastCase}, we can obtain that $A,B\in\mathcal{O}(n)$ satisfy \eqref{eq:lastCase} if and only if they satisfy
\begin{equation}\label{eq:BAdiag}
D_i(x) = \frac{k_i}{2} BA^\top.
\end{equation}
Since we look for a $D_i(x)$ independent of $A$, the right-hand side of the previous expression must be constant. This solution only exists by setting $A=PB$ with $P\in\{I_n,-I_n\}$, as we also derived in the previous steps of the proof.

Consider the case $P=I_n$ first. This reasoning implies that, for any $x\in\mathbb{R}^n$, $D_i(x)\in\mathcal{D}(k_i/2)$. We conclude that $\sigma_i:\mathbb{R}\to\mathbb{R}$ is affine and satisfies
\[
d_i\sigma_i(x)=d_i\frac{k_i}{2} x + \mu_i = -\frac{1}{\ell_i}(\ell_i^2-1+b_id_i^2)x+\mu_i,
\]
for a scalar $\mu_i\in\mathbb{R}$. On the other hand, if $P=-I_n$, we get $D_i(x)\in\mathcal{D}(-k_i/2)$ and
\[
\begin{aligned}
d_i\sigma_i(x) & = -d_i\frac{k_i}{2}x+\mu_i = \frac{1}{\ell_i}(\ell_i^2-1+b_id_i^2)x+\mu_i\\
& =(-d_i)\frac{\ell_i^2-1+b_i(-d_i)^2}{2(-d_i)\ell_i}x+\mu_i=d_i'\frac{k_i'}{2}x+\mu_i
\end{aligned}
\]
where $k_i' = -\frac{\ell_i^2-1+b_i(d_i')^2}{\ell_id_i'}$ and $d_i'=-d_i$, and hence $D_i(x)\in\mathcal{D}(k_i'/2)$ as desired. Hence, the case $P=I_n$ encodes all the options.

Since $(D_i(x))^2 = \frac{k^2_i}{4} I_n$, i.e., $b_i=k^2_i/4$, the definition of $k_i$ implies the following equation
\[
\frac{k^2_i d_i^2}{4} + \ell_i^2 + k_i d_i \ell_i = 1.
\]
This equation gives $\frac{k_i}{2} \in \{(\pm1 - \ell_i)/d_i\}$, which is coherent with the condition $D_i(x)\in\mathcal{D}((\pm 1 - \ell_i)/d_i)$ obtained for the case $A=PB$ and $\ell_i\neq 0$, but is just a particular case since here we get affine maps. We remark that for this choice of $\sigma_i$, the map $F$ restricted to $\Omega_i$ writes
\[
F_{\Omega_i}(x)=\ell_i x + c_i + (p_i - \ell_i)x + v_i = p_ix + c_i + \mu_iB^\top b
\]
for an arbitrary pair of scalars $\mu_i\in\mathbb{R}$ and $p_i\in\{\pm 1\}$. 


\section{Proof of Corollary \ref{co:corollary}}\label{app:proofCorollary}

We focus on a generic $\Omega_i\subset\mathbb{R}^n$, and consider
\[
G_{\Omega_i}(x) = \ell_i O x + c_i + d_i A^\top \sigma_i(Bx+b).
\]
Calling $y=y(x)=Ox$, we see that, by the orthogonality of $O$, it follows that
\[
G_{\Omega_i}(x)=\ell_i O x + c_i + d_i A^\top \sigma_i(BO^\top Ox+b) = \widehat{G}_{\Omega_i}(y(x)) 
\]
with 
\[
\widehat{G}_{\Omega_i}(y) = \ell_i y + c_i + d_i A^\top \sigma_i(BO^\top y + b).
\]
Recall that $G_{\Omega_i}$ has an orthogonal Jacobian matrix if and only if
\[
\begin{split}
G_{\Omega_i}'(x)(G_{\Omega_i}'(x))^\top = I_n &=(G_{\Omega_i}'(x))^\top G_{\Omega_i}'(x).
\end{split}
\]
Equivalently, by the chain rule, we get
\[
\begin{split}
I_n &=\widehat{G}_{\Omega_i}'(y(x))y'(x)y'(x)^\top (\widehat{G}_{\Omega_i}'(y(x)))^\top, \\
I_n &=y'(x)^\top (\widehat{G}_{\Omega_i}'(y(x)))^\top \widehat{G}_{\Omega_i}'(y(x))y'(x).
\end{split}
\]
Since $y'(x) = O$ is orthogonal, the two conditions are equivalent to $\widehat{G}_{\Omega_i}'(y)$ being orthogonal a.e.\,. Since $\widehat{B}:=BO^\top$ is orthogonal as well, $\widehat{G}_{\Omega_i}$ has the same form as case \ref{cii}, which leads to the condition $A = BO^\top$, as desired. Thus, we conclude that
\[
\begin{split}
G_{\Omega_i}(x) &= \ell_i O x + c_i + d_i O B^\top \sigma_i(Bx+b) \\
&= O(\ell_i x + c_i + d_i B^\top \sigma_i(Bx+b))=OF_{\Omega_i}(x)
\end{split}
\]
as desired.

\section{Further results on the limit layers}\label{app:limitLayers}

We start by providing a complete proof of Theorem \ref{thm:density}. This proof relies on a well-known characterization of piecewise affine functions.
Let us first introduce a generic characterization of the activation functions we are considering.
\begin{lemma}[Theorem 4.1 in \cite{ovchinnikov2000max}]\label{lemma:maxMin}
Let $\sigma:\mathbb{R}\to\mathbb{R}$ be a continuous piecewise affine function. Then, there exists a $k\in\mathbb{N}$, and a choice of scalars $a_{i,j},b_{i,j}\in\mathbb{R}$, $i=1,\cdots,k$, $j=1,\cdots,\ell_i$, such that
\begin{equation}\label{eq:pwl}
\sigma(x) = \max\{\sigma^1(x),\cdots,\sigma^k(x)\},
\end{equation}
where, for every $i =1, \cdots, k$, we set
\[
\sigma^i(x) = \min\{a_{i,1} x+b_{i,1},\cdots,a_{i,\ell_i}x+b_{i,\ell_i}\}.
\]
\end{lemma}

For an activation $\sigma\in\mathcal{L}(\alpha,\beta)$, $\alpha,\beta\in\mathbb{R}$, it is immediate to see that $a_{i,j}\in\{\alpha,\beta\}$ for every $i=1,\cdots,k$ and $j=1,\cdots,\ell_i$. In such a case, we recall the notation $\sigma(\cdot) = \sigma(\cdot\,; \alpha,\beta)$. 

\begin{proof}[Proof of Theorem \ref{thm:density}]
Assume $\Omega$ is compact, and consider an activation function $\sigma$ as in \eqref{eq:pwl} with slopes limited to a set of two elements. 
Fix $\varepsilon>0$ and two continuous functions $m,q:\Omega\to\mathbb{R}$. Call $M_{\Omega}:=\max_{x\in\Omega}\|x\|_2$ and consider a finite collection of sets $\{\Omega_1,\cdots,\Omega_N\}$ in $\mathcal{P}(\Omega)$ such that
\[
\begin{aligned}
\max_{x\in\Omega}|m(x) - \widetilde{m}(x)| &\leq \frac{\varepsilon}{4\sqrt{n}(M_{\Omega}\sqrt{n}+\|b\|_2)},\\
\max_{x\in\Omega}|q(x) - \widetilde{q}(x)| &\leq \frac{\varepsilon}{2},
\end{aligned}
\]
where $\widetilde{m},\widetilde{q}:\Omega\to\mathbb{R}$ are piecewise constant approximations of $m$ and $q$, which are constant on each of the sets $\Omega_1,\cdots,\Omega_N$.

Consider now the two functions $\mathcal{F}_{\Omega}(x;m,q)$ and $\mathcal{F}_{\Omega}(x;\widetilde{m},\widetilde{q})$
and write the difference
\[
\begin{split}
&\mathcal{F}_{\Omega}(x;m,q)-\mathcal{F}_\Omega(x;\widetilde{m},\widetilde{q})\\
& \qquad \qquad = (m(x)-\widetilde{m}(x))x + (q(x)-\widetilde{q}(x))\\
& \qquad \qquad +1_{\Omega}(x)B^\top\Bigl(\sigma(Bx+b;1-m(x),-1-m(x))\\
& \qquad \qquad - \sigma(Bx+b;1-\widetilde{m}(x),-1-\widetilde{m}(x))\Bigr).
\end{split}
\]
We now explicitly write the two functions whose difference needs to be controlled:
\[
\begin{split}
& \sigma(Bx+b;1-m(x),-1-m(x))= \max_{i=1,\cdots,k}\min_{j=1,\cdots,\ell_i}(a_{i,j}(x)(Bx+b) + b_{i,j}) \\
&\sigma(Bx+b;1-\widetilde{m}(x),-1-\widetilde{m}(x)) = \max_{i=1,\cdots,k}\min_{j=1,\cdots,\ell_i}(\widetilde{a}_{i,j}(x)(Bx+b) + {b}_{i,j}).
\end{split}
\]
Since the expression based on $\widetilde{m}$ and $\widetilde{q}$ can be constructed specifically for the target function at hand, we can assume that $a_{i,j}$ and $\widetilde{a}_{i,j}$ have agreeing expressions, meaning that if $a_{i,j}(x)=1-m(x)$, then $\widetilde{a}_{i,j}(x)=1-\widetilde{m}(x)$, for example. It follows that, for any $x\in\Omega$, one has
\[
\begin{split}
&\|\sigma(Bx+b;1-m(x),-1-m(x))-\sigma(Bx+b;1-\widetilde{m}(x),-1-\widetilde{m}(x))\|_\infty\\
&=\max_{r=1,\cdots,n} |\sigma((Bx+b)_r;1-m(x),-1-m(x))-\sigma((Bx+b)_r;1-\widetilde{m}(x),-1-\widetilde{m}(x))| \\
&\leq \max_{r=1,\cdots,n} \max_{i,j}|(a_{i,j}(x)-\widetilde{a}_{i,j}(x))(Bx+b)_r| = \max_{i,j} \max_{r=1,\cdots,n} |(a_{i,j}(x)-\widetilde{a}_{i,j}(x))(Bx+b)_r|\\
&= \max_{i,j} \|(a_{i,j}(x)-\widetilde{a}_{i,j}(x))(Bx+b)\|_\infty\leq \left(\max_{i,j} |a_{i,j}(x)-\widetilde{a}_{i,j}(x)|\right )\|Bx+b\|_\infty ,
\end{split}
\]
where $a_{i,j}(x)\in\{1-m(x),-1-m(x)\}$, $\widetilde{a}_{i,j}(x)\in\{1-\widetilde{m}(x),-1-\widetilde{m}(x)\}$, and $b_{i,j}\in\mathbb{R}$. This means that
\[
\begin{aligned}
\max_{i,j}|a_{i,j}(x)-\widetilde{a}_{i,j}(x)| & \leq |m(x)-\widetilde{m}(x)|\leq \frac{\varepsilon}{4\sqrt{n}(M_{\Omega}\sqrt{n}+\|b\|_2)}.
\end{aligned}
\]
Thus, it follows that
\[
\begin{aligned}
&\|\left(\sigma(Bx+b;1-m(x),-1-m(x)) -\sigma(Bx+b;1-\widetilde{m}(x),-1-\widetilde{m}(x))\right)\|_\infty\\
&\leq \sqrt{n} \frac{\varepsilon}{4\sqrt{n}(M_\Omega\sqrt{n}+\|b\|_2)}(M_\Omega\sqrt{n}+\|b\|_2) = \frac{\varepsilon}{4}.
\end{aligned}
\]

We then obtain the desired result
\[
\begin{aligned}
&\max_{x\in\Omega}\|\mathcal{F}_{\Omega}(x;m,q)-\mathcal{F}_\Omega(x;\widetilde{m},\widetilde{q})\|_\infty\leq \frac{\varepsilon}{4\sqrt{n}(M_\Omega\sqrt{n} + \|b\|_2)}M_\Omega + \frac{\varepsilon}{2} + \frac{\varepsilon}{4}\leq \frac{\varepsilon}{4} + \frac{\varepsilon}{2} + \frac{\varepsilon}{4} = \varepsilon
\end{aligned}
\]
as desired, where we used the estimate
\[
\begin{split}
\|(m(x)-\widetilde{m}(x))x\|_\infty &\leq \max_{x\in\Omega}|m(x)-\widetilde{m}(x)| \|x\|_2\leq \frac{\varepsilon}{4\sqrt{n}(M_\Omega\sqrt{n} + \|b\|_2)}M_\Omega\leq \frac{\varepsilon \sqrt{n}(M_\Omega\sqrt{n}+\|b\|_2)}{4\sqrt{n}(M_\Omega\sqrt{n} + \|b\|_2)} = \frac{\varepsilon}{4}.
\end{split}
\]
This allows us to conclude the proof.
\end{proof}

Theorem \ref{thm:density} opens up to a much broader class of network layers presenting non-affine skip connections that can be approximated arbitrarily well by choosing a fine enough partition $\{\Omega_1, \cdots, \Omega_N\}$ in $\mathcal{P}(\Omega)$.

Following up on this same idea, let us restrict our attention to the case $\sigma(x;\,\alpha,\beta) = \max\{\alpha x,\beta x\}$, with $\alpha,\beta\in\R$. Despite generally $x\mapsto \mathcal{F}_\Omega(x;m,q)$ does not have an orthogonal Jacobian, it provides a generalization of $x\mapsto x-2B^\top \mathrm{ReLU}(Bx+b)$, renaming $-B$ and $-b$ as $B$ and $b$, respectively, which is known to have an orthogonal Jacobian matrix a.e.\,. This argument suggests that maps of the form
\begin{align}
\mathcal{F}_\Omega(x;m,q) & = m(x)x+q(x) + (1-m(x))(x+B^\top b)+2B^\top \mathrm{ReLU}(-Bx-b)\label{eq:limit_layers}\\
&=q(x)+(1-m(x))B^\top b+\left( x + 2B^\top \mathrm{ReLU}(-Bx-b)\right),\nonumber
\end{align}
with $m,q:\Omega\to\R$ continuous functions, are potentially more expressive than maps of the type $x\mapsto x-2B^\top \mathrm{ReLU}(Bx+b)$. Then, in order to guarantee the trainability of these models, we can operate as follows.
\begin{theorem}\label{thm:dynIsometry}
Consider a convex and connected set $\Omega\subseteq\mathbb{R}^n$, a vector $b\in\mathbb{R}^n$, and two Lipschitz continuous functions $m,q:\Omega\to\mathbb{R}$ with Lipschitz constants $\mathrm{Lip}(m)$ and $\mathrm{Lip}(q)$. Let $\varepsilon > 0$ be such that
\[
\mathrm{Lip}(m)\|b\|_2 \leq \frac{\varepsilon}{2}, \text{ and } \mathrm{Lip}(q) \leq \frac{\varepsilon}{2\sqrt{n}}
\]
for a.e.\, $x\in\Omega$. Then, all the singular values of the Jacobian of $f(x)=\mathcal{F}_{\Omega}(x;m,q)$, when it is well defined, belong to the interval $[1-\varepsilon,1+\varepsilon]$.
\end{theorem}
\begin{proof}
Since $q$, $m$ are Lipschitz continuous, they are also differentiable a.e.\, by Rademacher's Theorem, see Theorem 3.1.6 \cite{federer2014geometric}. Hence, for a.e.\, $x \in \mathbb{R}^n$, we can compute the Jacobian of $f$ as
\begin{equation}\label{eq:limit_jac}
f'(x)= 1_n\nabla q(x)^\top - B^\top b\nabla m(x)^\top + (I_n- 2B^\top D(x)B),
\end{equation}
where $D(x)=\mathrm{diag}(\mathrm{ReLU}'(Bx+b))\in \{D=\mathrm{diag}(d_1,\cdots,d_n)\,|\,d_1,\cdots,d_n\in\{0, 1\}\}$. We know that $I_n-2B^\top D(x)B$ is orthogonal, so $f'(x)$ is a rank-two perturbation of an orthogonal matrix. When $m$ and $q$ are constant functions, we recover that $f'(x)$ is orthogonal as known from Theorem \ref{thm:char_RtoR_fields}.

By Weyl's inequality for singular values, see Corollary 7.3.5 in \cite{horn2012matrix}, we get that the $i$-th singular value of $f'(x)$, i.e., $\sigma_i(f'(x))$, satisfies
\begin{align*}
\left|\sigma_i\left(f'(x)\right)-1\right|&\leq \left\|1_n\nabla q(x)^\top - B^\top b \nabla m(x)^\top\right\|_2\\
&\leq \|b\|_2\|\nabla m\|_2 + \sqrt{n}\|\nabla q\|_2 \leq \mathrm{Lip}(m)\|b\|_2 + \sqrt{n}\mathrm{Lip}(q) \leq \varepsilon.
\end{align*}
We conclude that, for any $i\in\{1,\cdots,n\}$ and almost any $x\in\Omega$, $\sigma_i(f'(x))\in [1-\varepsilon,1+\varepsilon]$ holds true.
\end{proof}

The dynamical isometry principle requires the network Jacobian to have singular values in a neighbourhood of $1$. Thus, Theorem \ref{thm:dynIsometry} provides a practical sufficient way to promote this property over these limit networks by picking functions $m,q:\Omega\to\mathbb{R}$ with a moderate Lipschitz constant, and regularising for $\|b\|_2$ to be small. This must be done for every layer, so the whole network satisfies an analogous estimate. 
In agreement with Theorem \ref{thm:char_RtoR_fields}, we remark that the reasoning in Theorem \ref{thm:dynIsometry} could be repeated on each set of a partition $\{\Omega_1,\cdots,\Omega_N\}$ in $\mathcal{P}(\Omega)$, where we would recover the exact orthogonality for $m=\widetilde{m}$ and $q=\widetilde{q}$ constant on each subset $\Omega_i$ for the partition.

The bound provided by Theorem \ref{thm:dynIsometry}, however, does not investigate the actual effect of the perturbation on the singular values. We can refine our analysis as follows.
\begin{lemma}[Rank limited perturbation]\label{lemma:rankPerturbation}
Let $r,n\in\mathbb{N}$, $r<\lfloor n/2 \rfloor$, $M \in\mathbb{R}^{n\times n}$ be a rank r matrix, and $R\in\mathcal{O}(n)$. Then at least $n-2r$ singular values of $N=R+M$ are equal to $1$.
\end{lemma}
\begin{proof} By direct calculation, we see that
\[
N^\top N = I_n + R^\top M + M^\top R + M^\top M = U^\top (I_n + \Sigma_{2r})U,
\]
where $U^\top \Sigma_{2r}U$ is the diagonalization of the symmetric matrix $\Delta = R^\top M + M^\top R + M^\top M=R^\top M + M^\top(R+M)$, and $U^\top U = I_n$. The subscript $2r$ in $\Sigma_{2r}$ is used to remark that $\Delta$ has a rank of at most $2r$. This result is a consequence of two properties of the matrix rank: for arbitrary $A,B\in\mathbb{R}^{n\times n}$ we have (i) $\mathrm{rank}(AB)\leq \min\{\mathrm{rank}(A),\mathrm{rank}(B)\}$, and (ii) $\mathrm{rank}(A+B)\leq \mathrm{rank}(A)+\mathrm{rank}(B)$. This allows us to conclude that $I_n+\Sigma_{2r}$ has at most $n-2r$ diagonal entries different from $1$ as desired.
\end{proof}
\begin{corollary}
At most four singular values of the Jacobian of $f(x)=\mathcal{F}_\Omega(x;m,q)$ in Theorem \ref{thm:dynIsometry} are different from $1$.
\end{corollary}
\begin{proof}
The proof immediately follows by noticing that 
\[
f'(x) = 1_n\nabla q(x)^\top - B^\top b\nabla m(x)^\top + R(x),
\]
where $R(x)$ is the orthogonal Jacobian of $x+2B^\top \mathrm{ReLU}(-Bx+b)$. The matrix $1_n\nabla q(x)^\top - B^\top b\nabla m(x)^\top$ has at most rank $r=2$, and hence Lemma \ref{lemma:rankPerturbation} allows us to conclude.
\end{proof}
This result opens the way to more sophisticated designs that align the subspaces associated with the perturbed singular values, in the spirit of our newly introduced persistent subspace orthogonality. We leave these considerations for further studies.

Finally, we address the trainability of these limit models and show that they achieve results similar to those with exact Jacobian orthogonality. In these experiments, we do not control or penalize the Lipschitz constants of $m$ and $q$, but we initialize the bias term $b$ to have moderate entries.

We focus on classifying images from the Fashion-MNIST dataset and train models with varying numbers of layers $L \in \{10, 50, 200\}$. More details on the experimental setup can be found in Appendix \ref{app:experiments}. We use layers of the form \eqref{eq:limit_layers} with $q(x) = 0$ for all $x \in \mathbb{R}^{n \times n}$, and test three choices of $m: \mathbb{R}^{n\times n} \to \mathbb{R}$:
\begin{enumerate}
\item $m_1(x) = 1$, which is our baseline and gives layers of the form $x \mapsto x - 2B^\top \mathrm{ReLU}(Bx + b)$,
\item $m_2(x) = \mathrm{exp}(-\|x\|_F^2)/100$, where $\|\cdot\|_F$ is the Frobenius norm, and we have $\|\nabla m_2(x)\|_F\in \left[0, e^{-1/2} \sqrt{2}/100 \right]$,
\item $m_3(x) = W_1\mathrm{ReLU}(W_2x + w_3)$, with $W_1 \in \mathbb{R}^{1\times 1\times C\times 1}$ , $W_2 \in \mathbb{R}^{n \times n \times C \times C}$, $w_3\in\mathbb{R}^{1\times 1 \times C}$ trainable weights, with $x\in\mathbb{R}^{n\times n\times C}$, and the action of the kernels are realized through convolutions.
\end{enumerate}
The weights in $m_3$ and the bias vector $b$ are initialized from a random normal distribution of standard deviation $0.05$. Here, as activation function, we use $\sigma(x)=2\mathrm{ReLU}(x)$, in agreement with our theory in Section \ref{se:theory}. The convolutional weight $B$ is initialized with the Dirac initialization.

In Table \ref{tab:conv_limit_2}, we show that the limit models with $m_2$ and $m_3$, which we recall are not piecewise affine, present performances comparable to $m_1$, which is known to have an orthogonal Jacobian matrix a.e.\,. 
Because the norm $\|b\|_2$ is moderate at initialization, the trainability of these models is ensured by the fact that the Jacobian is close to orthogonal, given the estimates in Theorem \ref{thm:dynIsometry}.

\begin{table}[h!]
  \centering
  \caption{\textbf{Limit models.} The table reports the classification accuracy on the validation set of Fashion-MNIST. We test models with $L$ layers, $L\in\{10,50,200\}$. The first column indicates the function $m$ in Theorem \ref{thm:density}, whereas $q=0$. For $m_3$, we report the averages of two experiments, and the term after $\pm$ gives the interval where the results belong.}\label{tab:conv_limit_2}
  \vspace{10pt}
  \setlength{\tabcolsep}{10pt}                
  \begin{tabular}{c|ccc}
    \toprule
    $L$ & $\mathbf{m_{1}}$ & $\mathbf{m_{2}}$ & $\mathbf{m_{3}}$ \\   
    \midrule
    \textbf{10} & 87.2\% & 89.5\% & 89.6 $\pm$ 0.2\%\\
    \textbf{50} & 88.8\% & 89.4\% & 88.5 $\pm$ 0.4\%\\
    \textbf{200} & 88.2\% & 88.6\% & 87.9$\pm$ 0.1\%\\
    \bottomrule
  \end{tabular}
\end{table}

\section{Experimental setup}\label{app:experiments}
We provide additional details on the experimental setup for the experiments in Section \ref{se:experiments}, Section \ref{app:ablations} and Appendix \ref{app:limitLayers}. 

\paragraph{MNIST setup}
All models are trained with Adam on MNIST with standard normalization. We use the cross-entropy loss function. The purposes of the experiments that change $L$ and those that change $k$ differ, and so do their training setups. For the varying $L$ case, we consider learning rate $10^{-4}$, cosine annealing to $\eta_{\min}=5\times 10^{-6}$, batch size $256$, and $20$ epochs. This training setup is conservative in the learning rate since our goal is to find a good range where most models, even those with $L=200$ layers, don't exhibit immediate loss blowup. We report the results coming from the last model obtained at the end of the training procedure. In the experiments where we vary $k$, we instead fix $L=100$, which allows us to find a more optimized learning rate scheduler. We use the \textit{one cycle} scheduler, with percentage start $0.5$, minimum learning rate of $10^{-4}$ and maximum of $10^{-3}$.  We set a budget of $100$ epochs for each run, and early stopping occurs when the validation accuracy improves less than $0.5\%$ after a patience of $10$ epochs. We then report the results for the best model encountered during the training. For the experiment varying $L$ we initialize the weights of the first and last linear layers with PyTorch's default initialization, i.e., their entries are distributed as random uniform $\mathcal{U}([-\sqrt{\lambda},\sqrt{\lambda}])$ for $\lambda = 1/\mathrm{input\_features}$. For those varying $k$ we test the same initialization (in Section \ref{app:ablation_first_and_last}) as well as Stiefel initialization (in Section \ref{se:experiments}). All biases in the experiment with varying $L$ are initialized to zero, whereas those with varying $k$ are initialized to random uniform values, i.e., PyTorch's default. 

\paragraph{Fashion-MNIST setup}
The experiments in Appendix \ref{app:limitLayers} and in Section \ref{app:ablation_act_fn} are conducted on the Fashion-MNIST dataset. We use convolutional networks and minimise the cross-entropy loss function. Our theory targets layers that preserve the number of channels or the input dimension, like those in ResNets. When an expansion is needed, we add zero channels. After the convolutional layers we flatten the output and feed it to a trainable randomly initialized linear classifier. We set the number of channels to $d = 8$, the filter size to $3 \times 3$ and employ a Dirac initializer for the kernels and consider random normal biases. 
All models are trained using the Adam optimizer with a cosine decay learning rate scheduler. For the experiments with $50$ and $200$ layers we vary the learning rate in $[10^{-5},5\cdot 10^{-5}]$, whereas for those with $10$ layers in $[10^{-4},5\cdot 10^{-4}]$. We use batches of size $512$. Early stopping occurs when no improvement is observed within a sliding window of training epochs. We set patience to $10$ epochs, with a maximum number of epochs set to $400$. 

\paragraph{CIFAR-10 setup} In Section \ref{app:ablation_lr}, we train convolutional networks minimizing the cross-entropy loss to classify the CIFAR-10 dataset. We employ one initial layer augmenting the number of channels to $d = 16$ and we initialize its filters so that they act isometrically on the input. We fix the filter size to $3 \times 3$ and initialize the biases according to random normal distributions. At the end, we flatten the output and employ a linear classifier with weight initialized as a random Stiefel matrix. We do not use any learning rate scheduler so the starting learning rate is maintained throughout the training process.

\end{document}